\documentclass[11pt]{article}

\usepackage[utf8]{inputenc}
\usepackage{amsmath, amssymb}
\usepackage{graphicx}
\usepackage[table]{xcolor}  
\usepackage[T1]{fontenc}
\usepackage[utf8]{inputenc}

\usepackage{graphicx}
\usepackage{hyperref}
\usepackage{xcolor}
\usepackage[table]{xcolor}            
\definecolor{rowgray}{gray}{0.9} 
\usepackage{url}
\usepackage{booktabs}
\usepackage{float}
\usepackage{longtable}
\usepackage{lscape}
\usepackage{hyperref}
\usepackage{xcolor}
\usepackage{caption}
\usepackage{subcaption}
\usepackage{array}
\usepackage[margin=1in]{geometry}
\usepackage{cleveref}
\usepackage{libertine}
\usepackage{titlesec}
\usepackage{adjustbox}
\usepackage{comment}
\usepackage[authoryear]{natbib}

\title{The State of Large Language Models for African Languages: Progress and Challenges}
\author{
Kedir Yassin Hussen\textsuperscript{1}, 
Walelign Tewabe Sewunetie\textsuperscript{2}, 
Abinew Ali Ayele\textsuperscript{3}, 
Sukairaj Hafiz Imam\textsuperscript{4} \\
Eyob Nigussie Alemu\textsuperscript{5}, 
Shamsuddeen Hassan Muhammad\textsuperscript{6}, 
Seid Muhie Yimam\textsuperscript{7} \\
\\
\textsuperscript{1}University of Gondar, Ethiopia, 
\textsuperscript{2}AIMS Research and Innovation Centre, Rwanda \\
\textsuperscript{3}Bahir Dar University, Ethiopia, 
\textsuperscript{4}Bayero University Kano, India, 
\textsuperscript{5}Addis Ababa University, Ethiopia \\
\textsuperscript{6}Imperial College London, United Kingdom, 
\textsuperscript{7}University of Hamburg, Germany\\
Correspondence: \texttt{kediryassin25@yahoo.com}
}

\date{2025}
\begin{document}
\maketitle
\vspace{1em}
\begin{abstract}
Large Language Models (LLMs) are transforming Natural Language Processing (NLP), but their benefits are largely absent for Africa's 2,000 low-resource languages. This paper comparatively analyzes African language coverage across six LLMs, eight Small Language Models (SLMs), and six Specialized SLMs (SSLMs). The evaluation covers language coverage, training sets, technical limitations, script problems, and language modelling roadmaps. The work identifies 42 supported African languages and 23 available public data sets, and it shows a big gap where four languages (Amharic, Swahili, Afrikaans, and Malagasy) are always treated while there is over 98 \% of unsupported African languages. Moreover, the review shows that just Latin, Arabic, and Ge'ez scripts are identified while 20 active scripts are neglected. Some of the primary challenges are lack of data, tokenization biases, computational costs being very high, and evaluation issues. These issues demand language standardization, corpus development by the community, and effective adaptation methods for African languages.


\end{abstract}

\section{Introduction}
\label{sec:intro}

The rapid progress of Large Language Models (LLMs) has transformed the field of Natural Language Processing (NLP). However, these advancements have primarily concentrated on high-resource languages, leaving many low-resource languages, particularly African languages, largely overlooked. Africa has over 2,000 languages \citep{ethnologue}, the majority of which face significant challenges such as a lack of data, limited computational resources, insufficient NLP tools, and the absence of standardized benchmarks.

This study presents a three-stage review to evaluate LLMs' current status, challenges, and prospects for African languages. The first stage investigates both commercial and open-source LLMs models with more than 7 billion parameters regarding their support for African languages \citep{wang2024comprehensive}.
%
%
The second stage examines foundational multilingual models that have significantly influenced NLP research and development. Notably, these models include BERT \citep{devlin-etal-2019-bert}, mBERT \citep{wu-dredze-2020-languages}, T5 \citep{raffel2020exploring}, mT5 \citep{xue-etal-2021-mt5}, XLM \citep{lample2019cross}, RoBERTa \citep{liu2019roberta}, XLM-R \citep{conneau-etal-2020-unsupervised} and NLLB 200 \citep{nllb2022}. We refer to these models as Small Language Models (SLMs) due to their relatively smaller parameter counts compared to LLMs and their foundational role in the multilingual NLP ecosystem. These models were selected because they represent key milestones in multilingual transfer learning and remain widely used in academic and low-resource NLP research.
The third stage focuses on models specifically designed or fine-tuned for African languages. These include AfriBERTa \citep{ogueji2021afriberta}, AfriTeVa \citep{jude-ogundepo-etal-2022-afriteva}, AfroLM \citep{dossou-etal-2022-afrolm}, EthioLLM \citep{tonja-etal-2024-ethiollm}, EthioMT \citep{tonja-etal-2024-ethiomt}, and AfroXLMR \citep{alabi-etal-2022-adapting}. We call this category 'Specialised Small Language Models' (SSLMs) because they are generally based on SLM architectures but are adapted specifically to address the unique linguistic and structural properties of African languages. These models highlight recent efforts in African-centric NLP and address longstanding representational gaps.

The following objectives guide this review: 1) determine which African languages are most represented across LLMs, SLMs, and SSLMs, and analyze disparities in their coverage. 2) identify which African scripts face representational challenges in LLMs and why. 3) examine the technical limitations of representing African languages in LLMs, SLMs, and SSLMs. 4)review benchmark datasets and  models used in developing LLMs for African languages. 5) assess the future prospects of language modeling for African languages and outline the potential roadmap. This study seeks to explore various aspects of African language representation in LLMs, aiming to shed light on the current landscape and to contribute to the ongoing discussion of creating more inclusive and equitable NLP technologies.

\section{Related Work}

Many initiatives have concentrated on creating foundational monolingual models specifically designed for low-resource languages. These models are frequently developed from the ground up, utilizing well-established architectures like BERT and GPT. For instance, AraBERT was designed for various Arabic dialects and employs preprocessing techniques to normalize dialectal variations into Modern Standard Arabic \citep{antoun-etal-2020-arabert}. IndoBERT developed for the Indonesian language, adopts dynamic masking and sentence-piece tokenization to enhance linguistic representation \citep{koto-etal-2020-indolem}. Similarly, AraGPT2 integrates Arabic-specific tokenization to improve text generation quality, despite being trained on a relatively small 20GB dataset \citep{antoun-etal-2021-aragpt2}. Finnish GPT-2 demonstrates that domain-specific fine-tuning on smaller parameter models can outperform larger multilingual models, highlighting the effectiveness of focused, resource-efficient training approaches for underrepresented languages \citep{luukkonen-etal-2023-fingpt}.


Foundational monolingual models such as BERT, T5, DistilBERT \citep{sanh2019distilbert}, and BART \citep{lewis2019bart} have been adapted for multilingual tasks, despite being originally developed for high-resource languages. Adaptation methods include translate-train strategies, language adapters, and cross-lingual alignment. For example, BERT, although trained solely on English data, has served as the basis for multilingual variants such as mBERT, XLM-R, and LaBSE \citep{feng-etal-2022-language}. However, adapting monolingual models to multilingual low-resource settings presents several challenges. These include vocabulary and tokenization mismatches, insufficient pretraining data, representation misalignment, domain and script incompatibility, high computational costs, and limited evaluation resources. Such limitations highlight the complexities of extending monolingual architectures to linguistically diverse and underrepresented languages.


XLM-R, mBERT, mT5, BLOOM \citep{bloom-2022}, and mBART are foundational models trained with a multilingual corpus to be adapted to multilingual low-resource languages. XLM-R supports more than 100 languages, including low-resource languages. It combines RoBERTa and XLM and is pre-trained on 2.5TB using masked language modelling. It has limitations on script diversity and less performance language with fewer than 100k example sentences. mT5 is a text-to-text unified framework and supports 101 languages. It focuses on low-resource languages by training the model using mC4 datasets, which include 101 languages, many of which are low-resource languages.  

NLLB \citep{nllb-2022}, IndicBART \citep{dabre-etal-2022-indicbart}, AraT5 \citep{nagoudi-etal-2022-arat5}, Aya \citep{ustun-etal-2024-aya}, Glot500 \citep{imanigooghari-etal-2023-glot500} are mulitilingual languages designed for low resource languages. NLLB supports more than 200 languages with more than 7B parameters and is trained through human-in-the-loop data curation focused on low-resource languages with the limitation of data scarcity for extremely low-resource languages, computational cost, low-resource to low-resource translation underperformance, legal/religious text over-representation, and subword tokenization challenge. IndicBART is a BART-based multilingual model designed for 11 major Indian (low-resource) languages and uses separate BPE tokenizers for each script family. Aya 23 is a multilingual language model for 101 languages through instruction fine-tuning.  The model adapts the existing pretraining models like mT5 and BLOOM but focuses on human-annotated multilingual instruction datasets. 


Even though there is no clear distinction between SLMs and LLMs, \citet{lu-etal-2024-scaling} and \citet{wang2024comprehensive} provide some hints to categorise SLMs and LLMs. According to \citet{wang2024comprehensive}, 
LMs that have emergent ability are classified as LLMs, and LMs with the number of parameters less than 7B are classified as SLMs. In some cases, LLMs are impractical due to high computational demands or privacy concerns. \citet{wei2022chain} defines emergent ability as an ability to solve that is absent in smaller models, but present in LLMs. According to \citet{wang2024comprehensive}, all the models specialised for African languages are categorised as SLMs.

\section{\textbf{Methodology}}
This study employed a structured three-stage review methodology to examine the current status, challenges, and prospects of Language Models (LMs) for African languages. The review systematically analyzed a curated selection of models from three categories: (1) commercial and open source LLMs, (2) foundational Small Language Models (SLMs), and (3) Specialized Small Language Models (SSLMs) tailored for African languages.

\subsection{Model Selection}
We chose prominent and representative models in each category based on their visibility in the African NLP literature and support for low-resource languages. The details of the models are in Table \ref {tab:model_categories}.

\vspace{-10pt}
\begin{table}[htbp]
\small
\centering
\caption{Categorization of Language Models Reviewed}
\label{tab:model_categories}
\begin{tabular}{|l|p{3cm}|p{10cm}|}
\hline
\textbf{Category} & \textbf{Model Type} & \textbf{Examples} \\ \hline

\textbf{LLMs} & Large-Scale General-Purpose Models & 
GPT-4 \citep{openai2023gpt4}, Gemini 1.5 \citep{deepmind2023gemini}, PaLM \citep{palm2},
LLaMA 3 \citep{dubey2024llama}, DeepSeek V2 \citep{deepseekv3}, Aya 23 \citep{ustun-etal-2024-aya} \\ \hline

\textbf{SLMs} & Foundational Multilingual Models & 
BERT \citep{devlin-etal-2019-bert}, mBERT \citep{wu-dredze-2020-languages}, T5 \citep{raffel2020exploring}, mT5 \citep{xue-etal-2021-mt5}, XLM \citep{lample2019cross}, RoBERTa \citep{liu2019roberta}, XLM-R \citep{conneau-etal-2020-unsupervised}, NLLB 200 \citep{nllb2022} \\ \hline

\textbf{SSLMs} & Specialized African-Centric Models & 
AfriBERTa \citep{ogueji2021afriberta}, AfriTeVa \citep{jude-ogundepo-etal-2022-afriteva}, AfroLM \citep{dossou-etal-2022-afrolm}, EthioLLM \citep{tonja-etal-2024-ethiollm}, EthioMT \citep{tonja-etal-2024-ethiomt}, AfroXLMR \citep{alabi-etal-2022-adapting} \\ \hline

\end{tabular}
\end{table}
\vspace{-10pt}


\subsection{Review Procedure}

For each model, we reviewed official documentation, technical reports, and peer-reviewed publications to extract: (1) language and script coverage, especially for African languages; (2) tokenization strategies (e.g., BPE, SentencePiece, character-level encodings); (3) training objectives and corpora; and (4) model architecture, parameter size, and computational requirements.

\subsection{Dataset and Benchmark Mapping}
We mapped each model to African-relevant datasets and benchmarks to assess linguistic utility and task alignment (Appendix \ref{apd:second} Table \ref{tab:benchmarks}). We focused on datasets related to classification, named entity recognition, sentiment analysis, and machine translation. Models were evaluated based on their reported or inferable support for African languages and participation in benchmarks like MasakhaNER \citep{adelani2021masakhaner} and EthioBenchMarks\citep{tonja2024ethiollm}



\section{Discussion}
We use the information discussed, such as the dataset and architecture, to answer a series of questions about the status of LLMs in African languages.

\textbf{Question One.} \textit{Determining which African language is explored relatively more in LLMs, SLMs, and SSLMs and analysing any disparities in their coverage.}

\textbf{Answer.} Most of the LLMs like GPT-4, Gemini 1.5, PaLM 2, and DeepSeek have no clear documentation about the languages they support. This opacity makes it difficult to assess their true coverage and limits their accountability in addressing linguistic diversity. Foundational small language models, such as mBERT supports 104 languages, including 6 African languages \citep{bert_multilingual}. mT5 supports 101 languages, of which 14 are from Africa \citep{multilingual_t5}. XLM-R supports 100 languages, including 8 African languages. 

Although African languages have approximately 28.57\% of the 7,000 languages that exist around the world, underlying multilingual models considerably fail to represent them in proportion. For example, whereas mBERT for 104 languages should have approximately 30 African languages on board, it has only 6. Similarly, whereas mT5 should accommodate 29, it accommodates 14, and XLM-R has a mere 8 of a projected 29. This is indicative of the extreme underrepresentation of African languages in widely used language models and highlights the urgent need for more regionally and inclusively focused NLP efforts.

\begin{figure}[ht]
    \centering
    \begin{minipage}{0.48\textwidth}
        \centering
        \includegraphics[width=\linewidth]{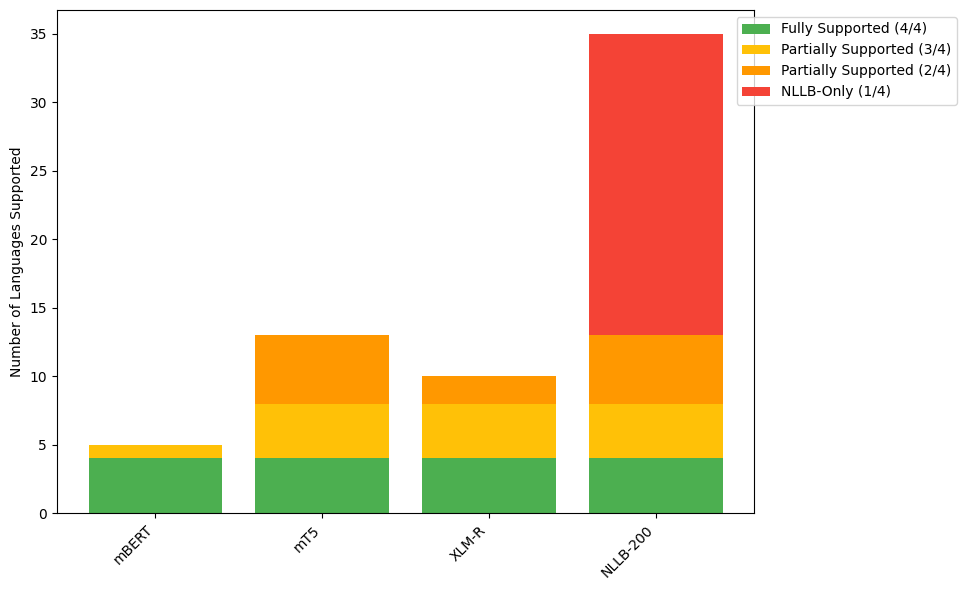}
        \caption{Languages  across SLMs.}
        \label{fig:plot}
    \end{minipage}\hfill
    \begin{minipage}{0.48\textwidth}
        All monolingual foundational  Small Language Models included in the study such as T5, BERT, RoBERTa do not support African languages directly, although some have been used as base models for further adaptations.
        Among the 38 African languages analyzed, only four Afrikaans, Amharic, Swahili, and Malagasy are fully supported by all multilingual SLMs.Appendix \ref{apd:second} and Table~\ref{tab:language_support} highlights the uneven distribution of African language support across SLMs and SSLMs, along with issues related to script compatibility and tokenization. 


    \end{minipage}
\end{figure}

NLLB-200-1.5B supports 200 languages, which include 38 African languages~\citep{nllb_200_distilled_600m}. As we can see from  \Cref{fig:plot} and Appendix \ref{apd:second} Table \ref{tab:foundation_models}, there are considerable disparities in the coverage of African languages in SLMs. We can observe that most SLMs cover only 38 languages out of 2000 languages in Africa. We can categorise SLMs into monolingual SLMs, such as BERT, T5, RoBERTa, and XLM and multilingual SLMs, which include mBERT, mT5, XLM-R and NLLB 200-1.5B.

 SSLMs show promise but face challenges like vocabulary, scripts, and tokenizer design, hindering equitable NLP development. This disparity underscores the urgent need for more inclusive AI development to bridge the linguistic gap and promote equitable access to technology across the continent. This uneven support is further illustrated in Figure~\ref{fig:plot}.



As shown in Table~\ref{tab:slm}, a total of 38 African languages are supported across six SLMs. Collectively, these models support approximately 42 African languages. A comprehensive list of the languages and their corresponding model support is provided in Appendix~\ref{apd:second}, Table~\ref{tab:language_support}.

Among the supported languages, the most represented language is Amharic, and it is due to relatively more digitized corpus availability, government and international dataset inclusion, and linguistic resource availability. Other languages supported include Somali, Swahili, and Yoruba, which are also supported based on wider regional usage, previous representation in multilingual contexts, and digital presence. Their frequent recurrence across many models shows not only linguistic expansion but also prior priority in the creation of materials, as opposed to merely their predominance on the continent.

\textbf{Question Two.} \textit{Which African scripts are getting challenged in the representation of LLMs and why?}

\textbf{Answer.}
There are approximately 37 writing systems historically used by African communities. Of these, 23 are currently in active use across different regions of the continent, while the remaining 14 are no longer in use \citep{worldwritingsystems}; see the Appendix \ref{apd:second} Table \ref{tab:scripts}. Among the active scripts, Latin, Ge’ez, and Arabic are the most widely used, collectively supporting 42 African languages.

According to the script, we can divide languages into non-script agnostic, partially script-agnostic, and fully script-agnostic languages \citep{conneau-etal-2020-unsupervised, xue-etal-2022-byt5}. Most African languages are challenged by non-script agnostic and partially script Agnostic models.  From 23 actively working scripts, only 3 are used in large language and small language models, which shows that script-wise African languages are not explored. Some language models, such as ByT5 \citep{xue-etal-2022-byt5} and CANINE \citep{clark-etal-2022-canine}, become fully script agnostic by avoiding script-specific tokenization and using a byte-level or character-level tokenizer. Large language models such as Gemini, GPT are partially script agnostic; they use sentence pieces and Byte Pair Encoding, which works well for different scripts, but they may fail to work on unseen scripts. 

The foundational SLMs like mT5, mBERT, XLM-R, and NLLB are partially script agnostic, and they work for different scripts, but they do not do script normalization, they don't do well on unseen scripts during the training. Specialized models such as AfriBERTa, EthioLLM, and EthioMT are not script-agnostic models while models like AfriTeVa, AfroML, and AfroXLM-R are partially script-agnostic models. Which implies they don't work well on unseen data on the training. 

\textbf{Question Three:} \textit{What are the technical challenges faced in representation across LLMs, SLMs, and SSLMs for African languages?}

\textbf{Answer.}We can see the technical challenges of African language models in terms of large language models, foundational small language models, and specialized models for African languages on one hand, and in terms of fine-tuned models for African languages derived from existing models, language adaptation fine-tuned models for African languages, and models developed from scratch for African languages on the other.

\begin{figure}[ht]
    \centering
    \noindent
    
    \begin{minipage}[t]{0.49\textwidth}
        \centering
        \small
        \captionof{table}{Model Tokens and Parameters (Chinchilla Scaling)}
        \label{tab:chinchilla}
        \rowcolors{2}{rowgray}{white}
        \begin{tabular}{@{}lrr@{}}
            \toprule
            Name & \#Tokens & \# of Parameters \\
            \midrule
            AfriBERTa    & 108,800,600     & 5,440,030       \\
            AfriTeVa     & 108,800,600     & 5,440,030       \\
            AfroLM       & 259,396,720     & 12,969,836      \\
            EthioLLM     & 299,512,427     & 14,975,621      \\
            EthioMT      & 5,845,000       & 292,250         \\
            AfroXLMR     & 760,000,000     & 38,000,000      \\
            \bottomrule
        \end{tabular}
    \end{minipage}\hfill
    \begin{minipage}[t]{0.47\textwidth}
        \vspace{0.5em} 
        Technical specifications of the large language models are presented in Table \ref{tab:llm_details}. Aya was trained on approximately 500 billion tokens, whereas the rest of the models such as GPT-4, Gemini 1.5, PaLM 2, and LLaMA 3 were trained with the over one trillion token datasets. As seen in Table \ref{tab:llm_details}, the general challenges of LLMs are computational, cost to train the model, which is unaffordable for African low-resource language researchers.
       
    \end{minipage}
\end{figure}
  Other challenges in training LLMs include data quality, as they require over a trillion tokens, which is difficult to obtain at that scale. Regarding low-resource languages, collecting such large amounts of data is challenging due to issues with data quality and bias. Additional challenges for low-resource languages include memory constraints, as models often use more than a trillion parameters during training. 

Hallucination and safety challenges are recent problems shown on LLMs on low-resource languages \citep{guerreiro-etal-2023-hallucinations, shen-etal-2024-language}. The main reason behind the problem is a lack of quality data. 

The main technical challenges of foundational SLMs include limited capacity, multilingual trade-off, and fine-tuning that needs domain-specific data  are related to the limited capacity they have because it is directly related to the number of parameters in the model, the size and quality of the training data. 

Training a model with monolingual data and fine-tuning the model with monolingual data produces better results because of the tokenization methods used \citep{rust-etal-2021-good}. Monolingual models have low problems related to tokenization because it has flexibility in producing tokens and can engage native-speaking experts to incorporate language-specific rules, such as tokenizing compound words and morphological splits.

\vspace{-10pt}
\begin{table}[H]
\centering
\scriptsize
\caption{Technical Details of Large Language Models}
\label{tab:llm_details}
\rowcolors{2}{rowgray}{white}
\begin{tabular}{>{\raggedright}p{1cm} >{\raggedright}p{1.2cm} p{1cm} p{1.5cm} p{1.6cm} p{1.2cm} p{2cm} p{2.5cm}}
\toprule
\textbf{Model} & \textbf{Architec-ture} & \textbf{Para-meters} & \textbf{Tokeniza-tion} & \textbf{Training Data} & \textbf{Comput. Cost} & \textbf{Training Objective} & \textbf{Key Features} \\
\midrule
GPT-4 & Decoder-only & ~1.8T & BPE & ~13T tokens & \$100M+ & Autoregressive LM & Multimodal, strong reasoning \\
Gemini Ultra  & Hybrid Enc-Dec & ~1.5T & SentencePiece & ~10T tokens & \$100M+ & Masked LM + Autoregressive & Multimodal (text, images, video) \\
PaLM 2 & Decoder-only & 540B & SentencePiece & 10T tokens & \$10M+  & Autoregressive & Text-only\\
LLaMA 3 (70B)  & Decoder-only & 70B & BPE variant & ~5T tokens & ~\$20M & Autoregressive LM & Open-weight, efficient \\
DeepSeek-V3  & Decoder-only & ~500B & BPE & ~8T tokens & ~\$50M & Autoregressive LM & 128K context, strong Chinese/English \\
Aya 23 & Decoder-only & 8B & Unigram & ~500B tokens & ~\$5M & Instruction Tuning & 101-language focused \\
\bottomrule
\end{tabular}
\end{table}
\textbf{Question Four.}\textit{What are the benchmark datasets and  models in developing LLMs for African languages?}

\begin{figure}[ht]
    \centering
    \begin{minipage}{0.48\textwidth}
        \centering
        \includegraphics[width=\linewidth]{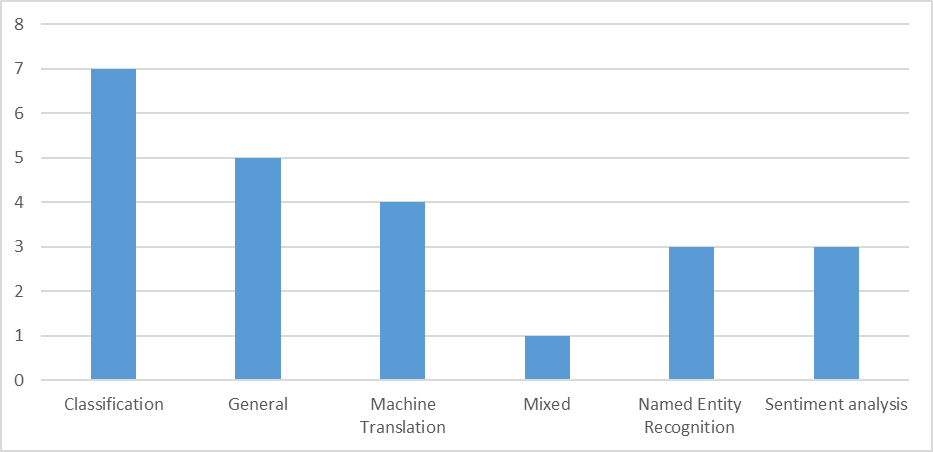}
        \caption{Different NLP tasks and the number of datasets prepared for the task}
        \label{fig:tasks}
    \end{minipage}\hfill
    \begin{minipage}{0.48\textwidth}
    
           \textbf{Question Five.}\textit{ Exploring the prospective of LLMs for Africa.} 
           
           \textbf{Answer.} Currently, artificial intelligence shows emergent ability which are arithmetic reasoning, agentic behaviour, common sense reasoning and symbolic reasoning. The path for this destination is very clear. Constructing LLMs for African languages is both challenge and an integral opportunity. Figure \ref{plot:prospect} outlines a strategic, step-by-step approach, beginning with foundational tasks such as language standardization, normalization, and formalization.
    \end{minipage}
\end{figure}

\textbf{Answer.} This study reveals that around 23 publicly available datasets are used by models solely SSLMs. Figure \ref{fig:tasks} shows the relationship between NLP tasks and the benchmark datasets prepared for those specific tasks. The study highlights disparities in the availability and utilization of benchmark datasets for African low-resource languages. Classification tasks have the highest number of datasets and models. This is likely because these tasks are simple and need less linguistic nuance than translation or named entity recognition. Appendix \ref{apd:second} Table \ref{tab:benchmarks} details the datasets benchmarks used across the specialised models for African languages. In the task column, General refers to a pretrain corpus, and mixed refers to a single benchmark which includes more than one task, like EthioBenchmark, which has five datasets for five tasks, including machine translation, part of speech tagging, classification, sentiment analysis, and named entity recognition.

 These are necessary for Africa, where linguistic diversity prevails, no orthographic consensus exists, and digital resources are scarce, hindering NLP development. Without rules for standard spelling and morphological annotation schemes, even basic text preprocessing becomes problematic.
\begin{figure}[H]
    \centering
    \includegraphics[width=0.8\linewidth]{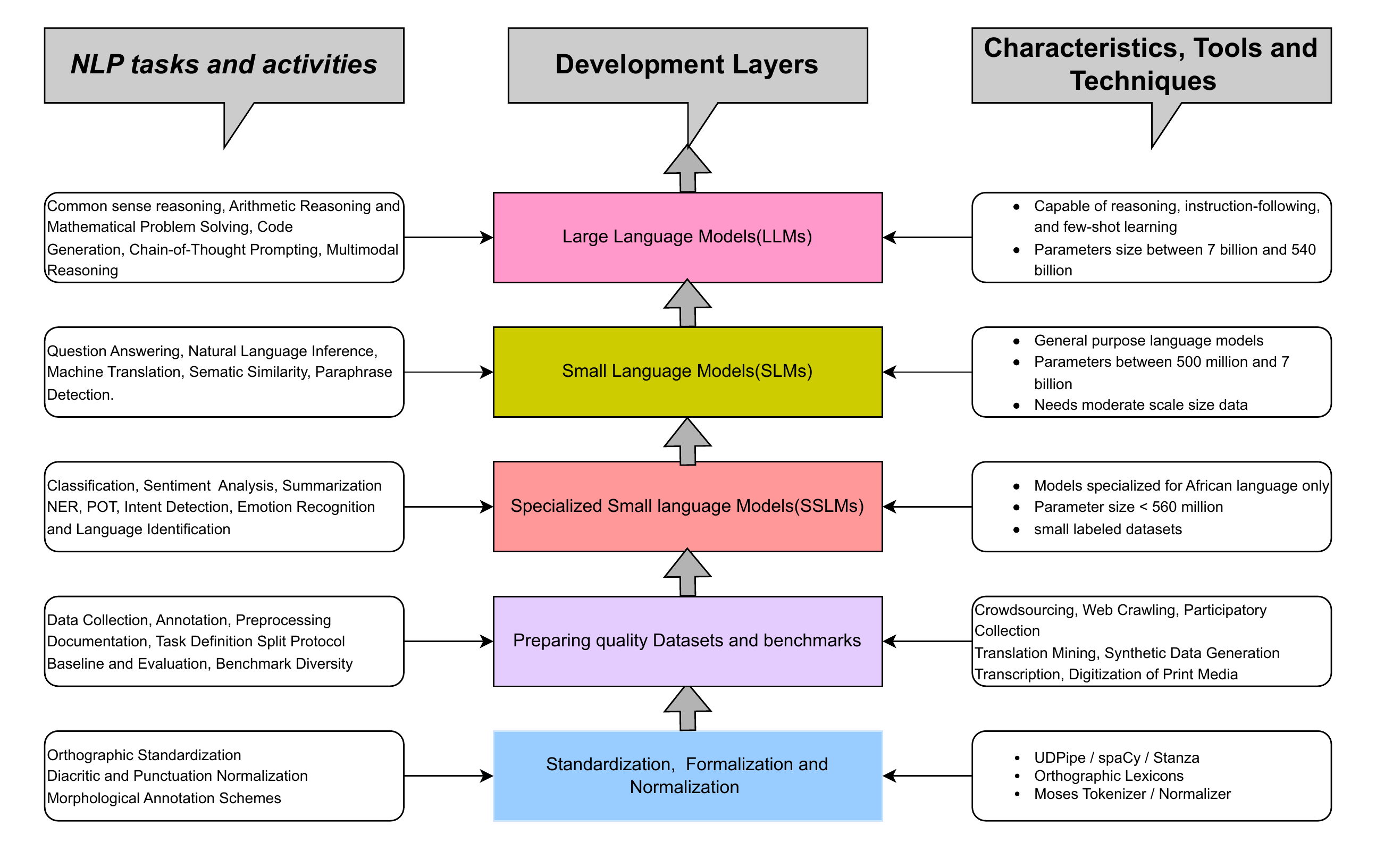}
    \caption{Figure 3: Roadmap for African Language-Model Development. The diagram proceeds bottom-
up. Foundational work on Standardisation, Formalisation, and Normalisation feeds directly
into Preparing Quality Datasets \& Benchmarks. These resources enable (i) SSLMs ($<$ 500 M
parameters) that target tightly scoped NLP tasks common in African contexts, (ii) General-
purpose SLMs (500 M–7 B parameters) capable of broader cross-lingual tasks, and finally (iii)
LLMs ($>$7 B parameters) that support advanced reasoning and generative capabilities. The left
column lists representative activities or tooling at each layer; the right column summarises the defining requirements. The central upward arrow highlights the dependency chain each layer builds on the assets and insights created in the layer(s) below.}
\label{plot:prospect}
\end{figure} The second level, preparation of dataset and benchmark, remains the main bottleneck. Most African languages are low-resource languages with few labeled data, small digitized corpora, and few benchmark resources. Crowdsourcing, participatory annotation, and digitization of oral and print sources are essential ways of bridging this gap. From these resources, the next step is to develop SSLMs that are task-specific and focused on individual languages with limited data. Such models serve as a stepping stone, enabling tangible NLP applications such as sentiment analysis and language identification and informing model construction.

SLMs of medium parameters enables multilingual capabilities and more universal tasks such as question answering and semantic similarity. Scaling to full-size LLMs, which require big data and computing capacity, remains an ambitious goal, given the continent's nascent AI infrastructure. Nonetheless, the roadmap shows a realistic and inclusive path—from foundational linguistic research to advanced models of reasoning—positioning Africa to ultimately solve its issues and build fair AI development for its multilingual populations. \citep{hoffmann2022training} states that, when the parameters of LLMs getting bigger, naturally it gets emergent ability and it can be unlock using chains of thought prompting. The paper claims when the parameters of LLMs reach around 540 billion the model naturally gain the emergent ability. 

 The chinchilla optimal ratio by \citet{hoffmann2022training} draws the relationship between the number of parameters used in the model with the number of tokens in the training datasets which is: 
 Number of Parameters (\(N\)) \(\approx 20 \times\) Number of Training Tokens (\(D\)).
 
 Let us assume specialised models for African languages trained with a quality corpus and datasets. Table \ref{tab:chinchilla} shows the number of parameters for each specialised model for African languages. The models are far too large to reach 540 billion parameters.

\begin{figure}[ht]
    \centering
    \includegraphics[width=1.0\textwidth, height=0.5\textwidth]{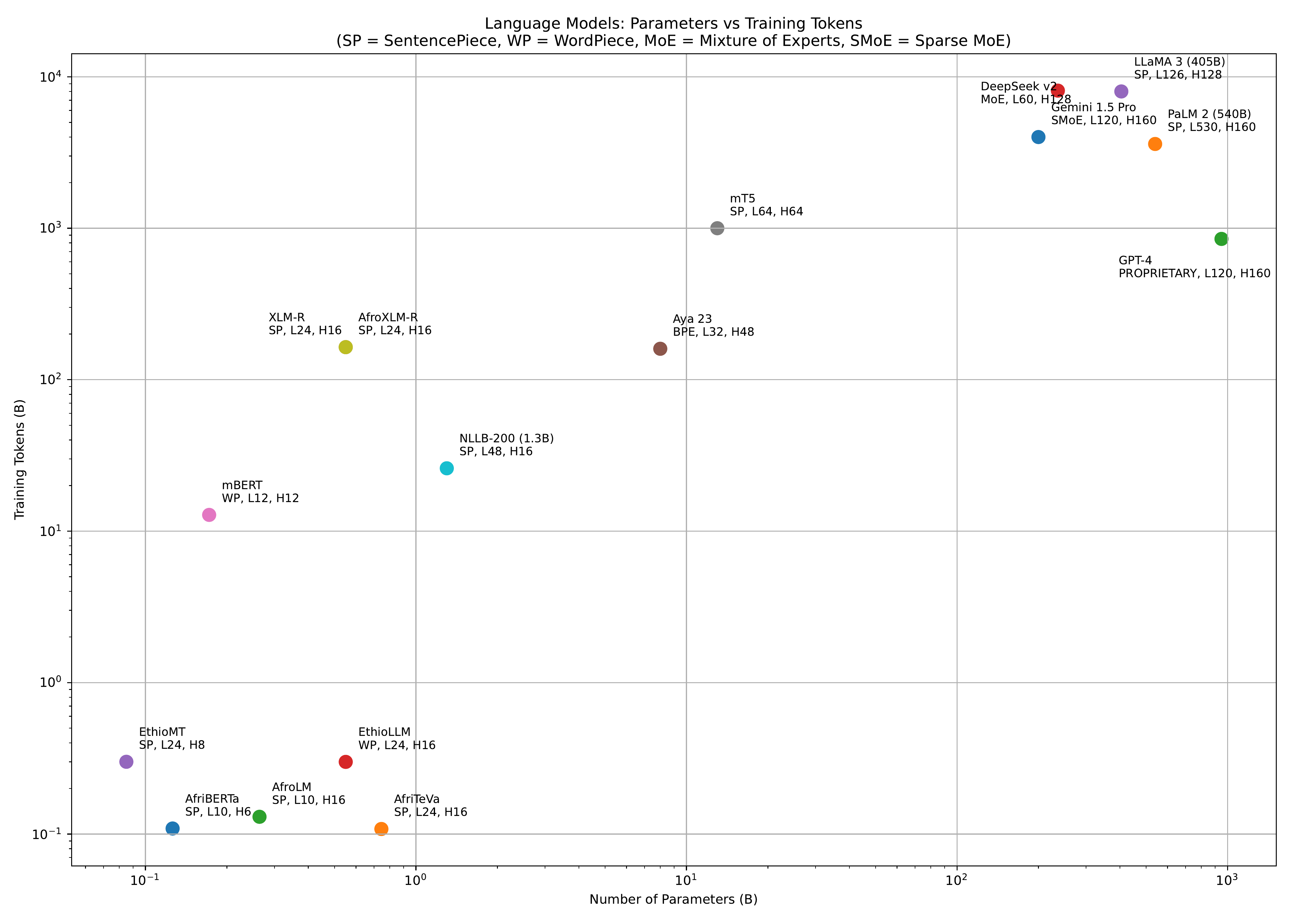}
    \caption{Model parameter size versus number of training tokens.}
    \label{fig:paravstoken}
\end{figure}

 As depicted in Figure \ref{fig:paravstoken}, models vary greatly in size, from small models like AfriBERTa (~10 million parameters) to large models like PaLM 2 (530B) and LLaMA 3 (405B). Larger models better capture linguistic nuances but need extensive resources. Most models use SentencePiece, while smaller ones use WordPiece. Large models are trained on billions of tokens for broader tasks; MoE architectures boost efficiency. Small models target specific languages, while large ones are multilingual, aiming for wider coverage and improved performance.
 
\vspace{-10pt}

\section*{Conclusion} This review reveals that African languages remain significantly underrepresented in current language models. Out of over 2,000 languages, only about 42 have any support in existing LLMs, SLMs, or SSLMs, primarily those with large speaker populations or official status. Script coverage is similarly limited, with just three scripts (Latin, Arabic, Ge'ez) being widely supported.

Major challenges include severe data scarcity, morphological complexity, a lack of standardized orthographies, and limited computational resources. Existing models, especially LLMs, require vast amounts of training data and infrastructure, posing substantial barriers to NLP development for African languages. Additionally, benchmark availability is sparse and unevenly distributed across tasks.

Despite these obstacles, progress with SSLMs shows potential for targeted advancement. A realistic roadmap begins with foundational linguistic work, followed by resource creation, and ultimately scalable models—offering a clear path toward inclusive language technologies for Africa.

 \section*{Recommendations} Advancing NLP for African languages requires developing tailored models like SSLMs and script-agnostic approaches, with a focus on improving data quality and culturally aware evaluations to reduce bias. It is also important to promote community-driven data collection, standardize scripts, and expand benchmarks across diverse tasks, while supporting open-access platforms. Additionally, securing institutional and government backing with funding, resources, and inclusion of African languages in digital services, alongside fostering international collaborations, will help elevate African language representation in AI research.

\bibliographystyle{plainnat}
\bibliography{pmlr-sample, anthology}

\begin{thebibliography}{48}
\providecommand{\natexlab}[1]{#1}
\providecommand{\url}[1]{\texttt{#1}}
\expandafter\ifx\csname urlstyle\endcsname\relax
  \providecommand{\doi}[1]{doi: #1}\else
  \providecommand{\doi}{doi: \begingroup \urlstyle{rm}\Url}\fi

\bibitem[Adelani et~al.(2021)Adelani, Abbott, Neubig, D’souza, Kreutzer, Lignos, Palen-Michel, Buzaaba, Rijhwani, Ruder, et~al.]{adelani2021masakhaner}
David~Ifeoluwa Adelani, Jade Abbott, Graham Neubig, Daniel D’souza, Julia Kreutzer, Constantine Lignos, Chester Palen-Michel, Happy Buzaaba, Shruti Rijhwani, Sebastian Ruder, et~al.
\newblock Masakhaner: Named entity recognition for african languages.
\newblock \emph{Transactions of the Association for Computational Linguistics}, 9:\penalty0 1116--1131, 2021.

\bibitem[Alabi et~al.(2022)Alabi, Adelani, Mosbach, and Klakow]{alabi-etal-2022-adapting}
Jesujoba~O. Alabi, David~Ifeoluwa Adelani, Marius Mosbach, and Dietrich Klakow.
\newblock Adapting pre-trained language models to {A}frican languages via multilingual adaptive fine-tuning.
\newblock In Nicoletta Calzolari, Chu-Ren Huang, Hansaem Kim, James Pustejovsky, Leo Wanner, Key-Sun Choi, Pum-Mo Ryu, Hsin-Hsi Chen, Lucia Donatelli, Heng Ji, Sadao Kurohashi, Patrizia Paggio, Nianwen Xue, Seokhwan Kim, Younggyun Hahm, Zhong He, Tony~Kyungil Lee, Enrico Santus, Francis Bond, and Seung-Hoon Na, editors, \emph{Proceedings of the 29th International Conference on Computational Linguistics}, pages 4336--4349, Gyeongju, Republic of Korea, October 2022. International Committee on Computational Linguistics.
\newblock URL \url{https://aclanthology.org/2022.coling-1.382/}.

\bibitem[Antoun et~al.(2020)Antoun, Baly, and Hajj]{antoun-etal-2020-arabert}
Wissam Antoun, Fady Baly, and Hazem Hajj.
\newblock {A}ra{BERT}: Transformer-based model for {A}rabic language understanding.
\newblock In Hend Al-Khalifa, Walid Magdy, Kareem Darwish, Tamer Elsayed, and Hamdy Mubarak, editors, \emph{Proceedings of the 4th Workshop on Open-Source Arabic Corpora and Processing Tools, with a Shared Task on Offensive Language Detection}, pages 9--15, Marseille, France, May 2020. European Language Resource Association.
\newblock ISBN 979-10-95546-51-1.
\newblock URL \url{https://aclanthology.org/2020.osact-1.2/}.

\bibitem[Antoun et~al.(2021)Antoun, Baly, and Hajj]{antoun-etal-2021-aragpt2}
Wissam Antoun, Fady Baly, and Hazem Hajj.
\newblock {A}ra{GPT}2: Pre-trained transformer for {A}rabic language generation.
\newblock In Nizar Habash, Houda Bouamor, Hazem Hajj, Walid Magdy, Wajdi Zaghouani, Fethi Bougares, Nadi Tomeh, Ibrahim Abu~Farha, and Samia Touileb, editors, \emph{Proceedings of the Sixth Arabic Natural Language Processing Workshop}, pages 196--207, Kyiv, Ukraine (Virtual), April 2021. Association for Computational Linguistics.
\newblock URL \url{https://aclanthology.org/2021.wanlp-1.21/}.

\bibitem[{Atelier National de Recherche Typographique (ANRT)} et~al.(2024){Atelier National de Recherche Typographique (ANRT)}, {Institut Designlabor Gutenberg (IDG)}, and {Script Encoding Initiative (SEI)}]{worldwritingsystems}
{Atelier National de Recherche Typographique (ANRT)}, {Institut Designlabor Gutenberg (IDG)}, and {Script Encoding Initiative (SEI)}.
\newblock The world's writing systems, 2024.
\newblock URL \url{https://www.worldswritingsystems.org/}.
\newblock Accessed: 2025-05-30.

\bibitem[Clark et~al.(2022)Clark, Garrette, Turc, and Wieting]{clark-etal-2022-canine}
Jonathan~H. Clark, Dan Garrette, Iulia Turc, and John Wieting.
\newblock Canine: Pre-training an efficient tokenization-free encoder for language representation.
\newblock \emph{Transactions of the Association for Computational Linguistics}, 10:\penalty0 73--91, 2022.
\newblock \doi{10.1162/tacl_a_00448}.
\newblock URL \url{https://aclanthology.org/2022.tacl-1.5/}.

\bibitem[Conneau et~al.(2020)Conneau, Khandelwal, Goyal, Chaudhary, Wenzek, Guzm{\'a}n, Grave, Ott, Zettlemoyer, and Stoyanov]{conneau-etal-2020-unsupervised}
Alexis Conneau, Kartikay Khandelwal, Naman Goyal, Vishrav Chaudhary, Guillaume Wenzek, Francisco Guzm{\'a}n, Edouard Grave, Myle Ott, Luke Zettlemoyer, and Veselin Stoyanov.
\newblock Unsupervised cross-lingual representation learning at scale.
\newblock In Dan Jurafsky, Joyce Chai, Natalie Schluter, and Joel Tetreault, editors, \emph{Proceedings of the 58th Annual Meeting of the Association for Computational Linguistics}, pages 8440--8451, Online, July 2020. Association for Computational Linguistics.
\newblock \doi{10.18653/v1/2020.acl-main.747}.
\newblock URL \url{https://aclanthology.org/2020.acl-main.747/}.

\bibitem[Costa-jussà et~al.(2022)Costa-jussà, Cross, Çelebi, Elbayad, Heafield, He, Kalbassi, Wang, Wang, Zhang, Fan, et~al.]{nllb2022}
Marta~R. Costa-jussà, James Cross, Onur Çelebi, Maha Elbayad, Kenneth Heafield, Yan He, Elahe Kalbassi, Linlu Wang, Long Wang, Shuo Zhang, Angela Fan, et~al.
\newblock No language left behind: Scaling human-centered machine translation.
\newblock In \emph{Proceedings of the 2022 Conference on Empirical Methods in Natural Language Processing (EMNLP)}, Abu Dhabi, UAE, 2022. Association for Computational Linguistics.

\bibitem[Dabre et~al.(2022)Dabre, Shrotriya, Kunchukuttan, Puduppully, Khapra, and Kumar]{dabre-etal-2022-indicbart}
Raj Dabre, Himani Shrotriya, Anoop Kunchukuttan, Ratish Puduppully, Mitesh Khapra, and Pratyush Kumar.
\newblock {I}ndic{BART}: A pre-trained model for indic natural language generation.
\newblock In Smaranda Muresan, Preslav Nakov, and Aline Villavicencio, editors, \emph{Findings of the Association for Computational Linguistics: ACL 2022}, pages 1849--1863, Dublin, Ireland, May 2022. Association for Computational Linguistics.
\newblock \doi{10.18653/v1/2022.findings-acl.145}.
\newblock URL \url{https://aclanthology.org/2022.findings-acl.145/}.

\bibitem[Dai et~al.(2023)Dai, Clark, Robinson, Moussalem, Ruder, Shakeri, Austin, et~al.]{palm2}
Andrew~M. Dai, Jonathan~H. Clark, Kevin Robinson, Maysam Moussalem, Sebastian Ruder, Siamak Shakeri, Jacob Austin, et~al.
\newblock Palm 2 technical report.
\newblock Technical report, Google, 2023.

\bibitem[DeepSeek-AI(2024)]{deepseekv3}
DeepSeek-AI.
\newblock Deepseek-v3 technical report.
\newblock arXiv preprint arXiv:2412.19437, December 2024.

\bibitem[Devlin et~al.(2019)Devlin, Chang, Lee, and Toutanova]{devlin-etal-2019-bert}
Jacob Devlin, Ming-Wei Chang, Kenton Lee, and Kristina Toutanova.
\newblock {BERT}: Pre-training of deep bidirectional transformers for language understanding.
\newblock In Jill Burstein, Christy Doran, and Thamar Solorio, editors, \emph{Proceedings of the 2019 Conference of the North {A}merican Chapter of the Association for Computational Linguistics: Human Language Technologies, Volume 1 (Long and Short Papers)}, pages 4171--4186, Minneapolis, Minnesota, June 2019. Association for Computational Linguistics.
\newblock \doi{10.18653/v1/N19-1423}.
\newblock URL \url{https://aclanthology.org/N19-1423/}.

\bibitem[Dossou et~al.(2022)Dossou, Tonja, Yousuf, Osei, Oppong, Shode, Awoyomi, and Emezue]{dossou-etal-2022-afrolm}
Bonaventure F.~P. Dossou, Atnafu~Lambebo Tonja, Oreen Yousuf, Salomey Osei, Abigail Oppong, Iyanuoluwa Shode, Oluwabusayo~Olufunke Awoyomi, and Chris Emezue.
\newblock {A}fro{LM}: A self-active learning-based multilingual pretrained language model for 23 {A}frican languages.
\newblock In Angela Fan, Iryna Gurevych, Yufang Hou, Zornitsa Kozareva, Sasha Luccioni, Nafise Sadat~Moosavi, Sujith Ravi, Gyuwan Kim, Roy Schwartz, and Andreas R{\"u}ckl{\'e}, editors, \emph{Proceedings of the Third Workshop on Simple and Efficient Natural Language Processing (SustaiNLP)}, pages 52--64, Abu Dhabi, United Arab Emirates (Hybrid), December 2022. Association for Computational Linguistics.
\newblock \doi{10.18653/v1/2022.sustainlp-1.11}.
\newblock URL \url{https://aclanthology.org/2022.sustainlp-1.11/}.

\bibitem[Dubey et~al.(2024)Dubey, Jauhri, Pandey, Kadian, Al-Dahle, Letman, Mathur, Schelten, Yang, Fan, Goyal, Hartshorn, Yang, Mitra, Sravankumar, Korenev, Hinsvark, Rao, Zhang, Rodriguez, Gregerson, Spataru, Roziere, Biron, Tang, Chern, Caucheteux, Nayak, Bi, Marra, McConnell, Keller, Touret, Wu, Wong, Ferrer, Nikolaidis, Allonsius, Song, Pintz, Livshits, Esiobu, Choudhary, Mahajan, Garcia-Olano, Perino, Hupkes, Lakomkin, AlBadawy, Lobanova, Dinan, Smith, Radenovic, Zhang, Synnaeve, Lee, Anderson, Nail, Mialon, Pang, Cucurell, Nguyen, Korevaar, Xu, Touvron, Zarov, Ibarra, Kloumann, Misra, Evtimov, Copet, Lee, Geffert, Vranes, Park, Mahadeokar, Shah, van~der Linde, Billock, Hong, Lee, Fu, Chi, Huang, Liu, Wang, Yu, Bitton, Spisak, Park, Rocca, Johnstun, Saxe, Jia, Alwala, Upasani, Plawiak, Li, Heafield, Stone, El-Arini, Iyer, Malik, Chiu, Bhalla, Rantala-Yeary, van~der Maaten, Chen, Tan, Jenkins, Martin, Madaan, Malo, Blecher, Landzaat, de~Oliveira, Muzzi, Pasupuleti, Singh, Paluri, Kardas, Oldham, Rita,
  Pavlova, Kambadur, Lewis, Si, Singh, Hassan, Goyal, Torabi, Bashlykov, Bogoychev, Chatterji, Duchenne, Çelebi, Alrassy, Zhang, Li, Vasic, Weng, Bhargava, Dubal, Krishnan, Koura, Xu, He, Dong, Srinivasan, Ganapathy, Calderer, Cabral, Stojnic, Raileanu, Girdhar, Patel, Sauvestre, Polidoro, Sumbaly, Taylor, Silva, Hou, Wang, Hosseini, Chennabasappa, Singh, Bell, Kim, Edunov, Nie, Narang, Raparthy, Shen, Wan, Bhosale, Zhang, Vandenhende, Batra, Whitman, Sootla, Collot, Gururangan, Borodinsky, Herman, Fowler, Sheasha, Georgiou, Scialom, Speckbacher, Mihaylov, Xiao, Karn, Goswami, Gupta, Ramanathan, Kerkez, Gonguet, Do, Vogeti, Petrovic, Chu, Xiong, Fu, Meers, Martinet, Wang, Tan, Xie, Jia, Wang, Goldschlag, Gaur, Babaei, Wen, Song, Zhang, Li, Mao, Coudert, Yan, Chen, Papakipos, Singh, Grattafiori, Jain, Kelsey, Shajnfeld, Gangidi, Victoria, Goldstand, Menon, Sharma, Boesenberg, Vaughan, Baevski, Feinstein, Kallet, Sangani, Yunus, Lupu, Alvarado, Caples, Gu, Ho, Poulton, Ryan, Ramchandani, Franco, Saraf,
  Chowdhury, Gabriel, Bharambe, Eisenman, Yazdan, James, Maurer, Leonhardi, Huang, Loyd, Paola, Paranjape, Liu, Wu, Ni, Hancock, Wasti, Spence, Stojkovic, Gamido, Montalvo, Parker, Burton, Mejia, Wang, Kim, Zhou, Hu, Chu, Cai, Tindal, Feichtenhofer, Civin, Beaty, Kreymer, Li, Wyatt, Adkins, Xu, Testuggine, David, Parikh, Liskovich, Foss, Wang, Le, Holland, Dowling, Jamil, Montgomery, Presani, Hahn, Wood, Brinkman, Arcaute, Dunbar, Smothers, Sun, Kreuk, Tian, Ozgenel, Caggioni, Guzmán, Kanayet, Seide, Florez, Schwarz, Badeer, Swee, Halpern, Thattai, Herman, Sizov, Guangyi, Zhang, Lakshminarayanan, Shojanazeri, Zou, Wang, Zha, Habeeb, Rudolph, Suk, Aspegren, Goldman, Damlaj, Molybog, Tufanov, Veliche, Gat, Weissman, Geboski, Kohli, Asher, Gaya, Marcus, Tang, Chan, Zhen, Reizenstein, Teboul, Zhong, Jin, Yang, Cummings, Carvill, Shepard, McPhie, Torres, Ginsburg, Wang, Wu, U, Saxena, Prasad, Khandelwal, Zand, Matosich, Veeraraghavan, Michelena, Li, Huang, Chawla, Lakhotia, Huang, Chen, Garg, A, Silva, Bell,
  Zhang, Guo, Yu, Moshkovich, Wehrstedt, Khabsa, Avalani, Bhatt, Tsimpoukelli, Mankus, Hasson, Lennie, Reso, Groshev, Naumov, Lathi, Keneally, Seltzer, Valko, Restrepo, Patel, Vyatskov, Samvelyan, Clark, Macey, Wang, Hermoso, Metanat, Rastegari, Bansal, Santhanam, Parks, White, Bawa, Singhal, Egebo, Usunier, Laptev, Dong, Zhang, Cheng, Chernoguz, Hart, Salpekar, Kalinli, Kent, Parekh, Saab, Balaji, Rittner, Bontrager, Roux, Dollar, Zvyagina, Ratanchandani, Yuvraj, Liang, Alao, Rodriguez, Ayub, Murthy, Nayani, Mitra, Li, Hogan, Battey, Wang, Maheswari, Howes, Rinott, Bondu, Datta, Chugh, Hunt, Dhillon, Sidorov, Pan, Verma, Yamamoto, Ramaswamy, Lindsay, Lindsay, Feng, Lin, Zha, Shankar, Zhang, Zhang, Wang, Agarwal, Sajuyigbe, Chintala, Max, Chen, Kehoe, Satterfield, Govindaprasad, Gupta, Cho, Virk, Subramanian, Choudhury, Goldman, Remez, Glaser, Best, Kohler, Robinson, Li, Zhang, Matthews, Chou, Shaked, Vontimitta, Ajayi, Montanez, Mohan, Kumar, Mangla, Albiero, Ionescu, Poenaru, Mihailescu, Ivanov, Li, Wang,
  Jiang, Bouaziz, Constable, Tang, Wang, Wu, Wang, Xia, Wu, Gao, Chen, Hu, Jia, Qi, Li, Zhang, Zhang, Adi, Nam, Yu, Wang, Hao, Qian, He, Rait, DeVito, Rosnbrick, Wen, Yang, and Zhao]{dubey2024llama}
Abhimanyu Dubey, Abhinav Jauhri, Abhinav Pandey, Abhishek Kadian, Ahmad Al-Dahle, Aiesha Letman, Akhil Mathur, Alan Schelten, Amy Yang, Angela Fan, Anirudh Goyal, Anthony Hartshorn, Aobo Yang, Archi Mitra, Archie Sravankumar, Artem Korenev, Arthur Hinsvark, Arun Rao, Aston Zhang, Aurelien Rodriguez, Austen Gregerson, Ava Spataru, Baptiste Roziere, Bethany Biron, Binh Tang, Bobbie Chern, Charlotte Caucheteux, Chaya Nayak, Chloe Bi, Chris Marra, Chris McConnell, Christian Keller, Christophe Touret, Chunyang Wu, Corinne Wong, Cristian~Canton Ferrer, Cyrus Nikolaidis, Damien Allonsius, Daniel Song, Danielle Pintz, Danny Livshits, David Esiobu, Dhruv Choudhary, Dhruv Mahajan, Diego Garcia-Olano, Diego Perino, Dieuwke Hupkes, Egor Lakomkin, Ehab AlBadawy, Elina Lobanova, Emily Dinan, Eric~Michael Smith, Filip Radenovic, Frank Zhang, Gabriel Synnaeve, Gabrielle Lee, Georgia~Lewis Anderson, Graeme Nail, Gregoire Mialon, Guan Pang, Guillem Cucurell, Hailey Nguyen, Hannah Korevaar, Hu~Xu, Hugo Touvron, Iliyan Zarov,
  Imanol~Arrieta Ibarra, Isabel Kloumann, Ishan Misra, Ivan Evtimov, Jade Copet, Jaewon Lee, Jan Geffert, Jana Vranes, Jason Park, Jay Mahadeokar, Jeet Shah, Jelmer van~der Linde, Jennifer Billock, Jenny Hong, Jenya Lee, Jeremy Fu, Jianfeng Chi, Jianyu Huang, Jiawen Liu, Jie Wang, Jiecao Yu, Joanna Bitton, Joe Spisak, Jongsoo Park, Joseph Rocca, Joshua Johnstun, Joshua Saxe, Junteng Jia, Kalyan~Vasuden Alwala, Kartikeya Upasani, Kate Plawiak, Ke~Li, Kenneth Heafield, Kevin Stone, Khalid El-Arini, Krithika Iyer, Kshitiz Malik, Kuenley Chiu, Kunal Bhalla, Lauren Rantala-Yeary, Laurens van~der Maaten, Lawrence Chen, Liang Tan, Liz Jenkins, Louis Martin, Lovish Madaan, Lubo Malo, Lukas Blecher, Lukas Landzaat, Luke de~Oliveira, Madeline Muzzi, Mahesh Pasupuleti, Mannat Singh, Manohar Paluri, Marcin Kardas, Mathew Oldham, Mathieu Rita, Maya Pavlova, Melanie Kambadur, Mike Lewis, Min Si, Mitesh~Kumar Singh, Mona Hassan, Naman Goyal, Narjes Torabi, Nikolay Bashlykov, Nikolay Bogoychev, Niladri Chatterji, Olivier
  Duchenne, Onur Çelebi, Patrick Alrassy, Pengchuan Zhang, Pengwei Li, Petar Vasic, Peter Weng, Prajjwal Bhargava, Pratik Dubal, Praveen Krishnan, Punit~Singh Koura, Puxin Xu, Qing He, Qingxiao Dong, Ragavan Srinivasan, Raj Ganapathy, Ramon Calderer, Ricardo~Silveira Cabral, Robert Stojnic, Roberta Raileanu, Rohit Girdhar, Rohit Patel, Romain Sauvestre, Ronnie Polidoro, Roshan Sumbaly, Ross Taylor, Ruan Silva, Rui Hou, Rui Wang, Saghar Hosseini, Sahana Chennabasappa, Sanjay Singh, Sean Bell, Seohyun~Sonia Kim, Sergey Edunov, Shaoliang Nie, Sharan Narang, Sharath Raparthy, Sheng Shen, Shengye Wan, Shruti Bhosale, Shun Zhang, Simon Vandenhende, Soumya Batra, Spencer Whitman, Sten Sootla, Stephane Collot, Suchin Gururangan, Sydney Borodinsky, Tamar Herman, Tara Fowler, Tarek Sheasha, Thomas Georgiou, Thomas Scialom, Tobias Speckbacher, Todor Mihaylov, Tong Xiao, Ujjwal Karn, Vedanuj Goswami, Vibhor Gupta, Vignesh Ramanathan, Viktor Kerkez, Vincent Gonguet, Virginie Do, Vish Vogeti, Vladan Petrovic, Weiwei Chu,
  Wenhan Xiong, Wenyin Fu, Whitney Meers, Xavier Martinet, Xiaodong Wang, Xiaoqing~Ellen Tan, Xinfeng Xie, Xuchao Jia, Xuewei Wang, Yaelle Goldschlag, Yashesh Gaur, Yasmine Babaei, Yi~Wen, Yiwen Song, Yuchen Zhang, Yue Li, Yuning Mao, Zacharie~Delpierre Coudert, Zheng Yan, Zhengxing Chen, Zoe Papakipos, Aaditya Singh, Aaron Grattafiori, Abha Jain, Adam Kelsey, Adam Shajnfeld, Adithya Gangidi, Adolfo Victoria, Ahuva Goldstand, Ajay Menon, Ajay Sharma, Alex Boesenberg, Alex Vaughan, Alexei Baevski, Allie Feinstein, Amanda Kallet, Amit Sangani, Anam Yunus, Andrei Lupu, Andres Alvarado, Andrew Caples, Andrew Gu, Andrew Ho, Andrew Poulton, Andrew Ryan, Ankit Ramchandani, Annie Franco, Aparajita Saraf, Arkabandhu Chowdhury, Ashley Gabriel, Ashwin Bharambe, Assaf Eisenman, Azadeh Yazdan, Beau James, Ben Maurer, Benjamin Leonhardi, Bernie Huang, Beth Loyd, Beto~De Paola, Bhargavi Paranjape, Bing Liu, Bo~Wu, Boyu Ni, Braden Hancock, Bram Wasti, Brandon Spence, Brani Stojkovic, Brian Gamido, Britt Montalvo, Carl
  Parker, Carly Burton, Catalina Mejia, Changhan Wang, Changkyu Kim, Chao Zhou, Chester Hu, Ching-Hsiang Chu, Chris Cai, Chris Tindal, Christoph Feichtenhofer, Damon Civin, Dana Beaty, Daniel Kreymer, Daniel Li, Danny Wyatt, David Adkins, David Xu, Davide Testuggine, Delia David, Devi Parikh, Diana Liskovich, Didem Foss, Dingkang Wang, Duc Le, Dustin Holland, Edward Dowling, Eissa Jamil, Elaine Montgomery, Eleonora Presani, Emily Hahn, Emily Wood, Erik Brinkman, Esteban Arcaute, Evan Dunbar, Evan Smothers, Fei Sun, Felix Kreuk, Feng Tian, Firat Ozgenel, Francesco Caggioni, Francisco Guzmán, Frank Kanayet, Frank Seide, Gabriela~Medina Florez, Gabriella Schwarz, Gada Badeer, Georgia Swee, Gil Halpern, Govind Thattai, Grant Herman, Grigory Sizov, Guangyi, Zhang, Guna Lakshminarayanan, Hamid Shojanazeri, Han Zou, Hannah Wang, Hanwen Zha, Haroun Habeeb, Harrison Rudolph, Helen Suk, Henry Aspegren, Hunter Goldman, Ibrahim Damlaj, Igor Molybog, Igor Tufanov, Irina-Elena Veliche, Itai Gat, Jake Weissman, James
  Geboski, James Kohli, Japhet Asher, Jean-Baptiste Gaya, Jeff Marcus, Jeff Tang, Jennifer Chan, Jenny Zhen, Jeremy Reizenstein, Jeremy Teboul, Jessica Zhong, Jian Jin, Jingyi Yang, Joe Cummings, Jon Carvill, Jon Shepard, Jonathan McPhie, Jonathan Torres, Josh Ginsburg, Junjie Wang, Kai Wu, Kam~Hou U, Karan Saxena, Karthik Prasad, Kartikay Khandelwal, Katayoun Zand, Kathy Matosich, Kaushik Veeraraghavan, Kelly Michelena, Keqian Li, Kun Huang, Kunal Chawla, Kushal Lakhotia, Kyle Huang, Lailin Chen, Lakshya Garg, Lavender A, Leandro Silva, Lee Bell, Lei Zhang, Liangpeng Guo, Licheng Yu, Liron Moshkovich, Luca Wehrstedt, Madian Khabsa, Manav Avalani, Manish Bhatt, Maria Tsimpoukelli, Martynas Mankus, Matan Hasson, Matthew Lennie, Matthias Reso, Maxim Groshev, Maxim Naumov, Maya Lathi, Meghan Keneally, Michael~L. Seltzer, Michal Valko, Michelle Restrepo, Mihir Patel, Mik Vyatskov, Mikayel Samvelyan, Mike Clark, Mike Macey, Mike Wang, Miquel~Jubert Hermoso, Mo~Metanat, Mohammad Rastegari, Munish Bansal, Nandhini
  Santhanam, Natascha Parks, Natasha White, Navyata Bawa, Nayan Singhal, Nick Egebo, Nicolas Usunier, Nikolay~Pavlovich Laptev, Ning Dong, Ning Zhang, Norman Cheng, Oleg Chernoguz, Olivia Hart, Omkar Salpekar, Ozlem Kalinli, Parkin Kent, Parth Parekh, Paul Saab, Pavan Balaji, Pedro Rittner, Philip Bontrager, Pierre Roux, Piotr Dollar, Polina Zvyagina, Prashant Ratanchandani, Pritish Yuvraj, Qian Liang, Rachad Alao, Rachel Rodriguez, Rafi Ayub, Raghotham Murthy, Raghu Nayani, Rahul Mitra, Raymond Li, Rebekkah Hogan, Robin Battey, Rocky Wang, Rohan Maheswari, Russ Howes, Ruty Rinott, Sai~Jayesh Bondu, Samyak Datta, Sara Chugh, Sara Hunt, Sargun Dhillon, Sasha Sidorov, Satadru Pan, Saurabh Verma, Seiji Yamamoto, Sharadh Ramaswamy, Shaun Lindsay, Shaun Lindsay, Sheng Feng, Shenghao Lin, Shengxin~Cindy Zha, Shiva Shankar, Shuqiang Zhang, Shuqiang Zhang, Sinong Wang, Sneha Agarwal, Soji Sajuyigbe, Soumith Chintala, Stephanie Max, Stephen Chen, Steve Kehoe, Steve Satterfield, Sudarshan Govindaprasad, Sumit Gupta,
  Sungmin Cho, Sunny Virk, Suraj Subramanian, Sy~Choudhury, Sydney Goldman, Tal Remez, Tamar Glaser, Tamara Best, Thilo Kohler, Thomas Robinson, Tianhe Li, Tianjun Zhang, Tim Matthews, Timothy Chou, Tzook Shaked, Varun Vontimitta, Victoria Ajayi, Victoria Montanez, Vijai Mohan, Vinay~Satish Kumar, Vishal Mangla, Vítor Albiero, Vlad Ionescu, Vlad Poenaru, Vlad~Tiberiu Mihailescu, Vladimir Ivanov, Wei Li, Wenchen Wang, Wenwen Jiang, Wes Bouaziz, Will Constable, Xiaocheng Tang, Xiaofang Wang, Xiaojian Wu, Xiaolan Wang, Xide Xia, Xilun Wu, Xinbo Gao, Yanjun Chen, Ye~Hu, Ye~Jia, Ye~Qi, Yenda Li, Yilin Zhang, Ying Zhang, Yossi Adi, Youngjin Nam, Yu, Wang, Yuchen Hao, Yundi Qian, Yuzi He, Zach Rait, Zachary DeVito, Zef Rosnbrick, Zhaoduo Wen, Zhenyu Yang, and Zhiwei Zhao.
\newblock The llama 3 herd of models, 2024.

\bibitem[{Ethnologue}(2025)]{ethnologue}
{Ethnologue}.
\newblock Ethnologue: Languages of the world, 2025.
\newblock Accessed: 2025-05-20.

\bibitem[Feng et~al.(2022)Feng, Yang, Cer, Arivazhagan, and Wang]{feng-etal-2022-language}
Fangxiaoyu Feng, Yinfei Yang, Daniel Cer, Naveen Arivazhagan, and Wei Wang.
\newblock Language-agnostic {BERT} sentence embedding.
\newblock In Smaranda Muresan, Preslav Nakov, and Aline Villavicencio, editors, \emph{Proceedings of the 60th Annual Meeting of the Association for Computational Linguistics (Volume 1: Long Papers)}, pages 878--891, Dublin, Ireland, May 2022. Association for Computational Linguistics.
\newblock \doi{10.18653/v1/2022.acl-long.62}.
\newblock URL \url{https://aclanthology.org/2022.acl-long.62/}.

\bibitem[Guerreiro et~al.(2023)Guerreiro, Alves, Waldendorf, Haddow, Birch, Colombo, and Martins]{guerreiro-etal-2023-hallucinations}
Nuno~M. Guerreiro, Duarte~M. Alves, Jonas Waldendorf, Barry Haddow, Alexandra Birch, Pierre Colombo, and Andr{\'e} F.~T. Martins.
\newblock Hallucinations in large multilingual translation models.
\newblock \emph{Transactions of the Association for Computational Linguistics}, 11:\penalty0 1500--1517, 2023.
\newblock \doi{10.1162/tacl_a_00615}.
\newblock URL \url{https://aclanthology.org/2023.tacl-1.85/}.

\bibitem[Hoffmann et~al.(2022)Hoffmann, Borgeaud, Mensch, Buchatskaya, Cai, Rutherford, de~Las~Casas, Hendricks, Welbl, Clark, et~al.]{hoffmann2022training}
Jordan Hoffmann, Sebastian Borgeaud, Arthur Mensch, Elena Buchatskaya, Trevor Cai, Eliza Rutherford, Diego de~Las~Casas, Lisa~Anne Hendricks, Johannes Welbl, Aidan Clark, et~al.
\newblock Training compute-optimal large language models.
\newblock \emph{arXiv preprint arXiv:2203.15556}, 2022.

\bibitem[Imani et~al.(2023)Imani, Lin, Kargaran, Severini, Jalili~Sabet, Kassner, Ma, Schmid, Martins, Yvon, and Sch{\"u}tze]{imanigooghari-etal-2023-glot500}
Ayyoob Imani, Peiqin Lin, Amir~Hossein Kargaran, Silvia Severini, Masoud Jalili~Sabet, Nora Kassner, Chunlan Ma, Helmut Schmid, Andr{\'e} Martins, Fran{\c{c}}ois Yvon, and Hinrich Sch{\"u}tze.
\newblock Glot500: Scaling multilingual corpora and language models to 500 languages.
\newblock In Anna Rogers, Jordan Boyd-Graber, and Naoaki Okazaki, editors, \emph{Proceedings of the 61st Annual Meeting of the Association for Computational Linguistics (Volume 1: Long Papers)}, pages 1082--1117, Toronto, Canada, July 2023. Association for Computational Linguistics.
\newblock \doi{10.18653/v1/2023.acl-long.61}.
\newblock URL \url{https://aclanthology.org/2023.acl-long.61/}.

\bibitem[Jude~Ogundepo et~al.(2022)Jude~Ogundepo, Oladipo, Adeyemi, Ogueji, and Lin]{jude-ogundepo-etal-2022-afriteva}
Odunayo Jude~Ogundepo, Akintunde Oladipo, Mofetoluwa Adeyemi, Kelechi Ogueji, and Jimmy Lin.
\newblock {A}fri{T}e{VA}: Extending ?small data? pretraining approaches to sequence-to-sequence models.
\newblock In Colin Cherry, Angela Fan, George Foster, Gholamreza~(Reza) Haffari, Shahram Khadivi, Nanyun~(Violet) Peng, Xiang Ren, Ehsan Shareghi, and Swabha Swayamdipta, editors, \emph{Proceedings of the Third Workshop on Deep Learning for Low-Resource Natural Language Processing}, pages 126--135, Hybrid, July 2022. Association for Computational Linguistics.
\newblock \doi{10.18653/v1/2022.deeplo-1.14}.
\newblock URL \url{https://aclanthology.org/2022.deeplo-1.14/}.

\bibitem[Koto et~al.(2020)Koto, Rahimi, Lau, and Baldwin]{koto-etal-2020-indolem}
Fajri Koto, Afshin Rahimi, Jey~Han Lau, and Timothy Baldwin.
\newblock {I}ndo{LEM} and {I}ndo{BERT}: A benchmark dataset and pre-trained language model for {I}ndonesian {NLP}.
\newblock In Donia Scott, Nuria Bel, and Chengqing Zong, editors, \emph{Proceedings of the 28th International Conference on Computational Linguistics}, pages 757--770, Barcelona, Spain (Online), December 2020. International Committee on Computational Linguistics.
\newblock \doi{10.18653/v1/2020.coling-main.66}.
\newblock URL \url{https://aclanthology.org/2020.coling-main.66/}.

\bibitem[Lample and Conneau(2019)]{lample2019cross}
Guillaume Lample and Alexis Conneau.
\newblock Cross-lingual language model pretraining.
\newblock \emph{Advances in Neural Information Processing Systems (NeurIPS)}, 32, 2019.

\bibitem[Lewis et~al.(2019)Lewis, Liu, Goyal, Ghazvininejad, Mohamed, Levy, Stoyanov, and Zettlemoyer]{lewis2019bart}
Mike Lewis, Yinhan Liu, Naman Goyal, Marjan Ghazvininejad, Abdelrahman Mohamed, Omer Levy, Ves Stoyanov, and Luke Zettlemoyer.
\newblock Bart: Denoising sequence-to-sequence pre-training for natural language generation, translation, and comprehension.
\newblock \emph{arXiv preprint arXiv:1910.13461}, 2019.

\bibitem[Liu et~al.(2019)Liu, Ott, Goyal, Du, Joshi, Chen, Levy, Lewis, Zettlemoyer, and Stoyanov]{liu2019roberta}
Yinhan Liu, Myle Ott, Naman Goyal, Jingfei Du, Mandar Joshi, Danqi Chen, Omer Levy, Mike Lewis, Luke Zettlemoyer, and Veselin Stoyanov.
\newblock Roberta: A robustly optimized bert pretraining approach.
\newblock \emph{arXiv preprint arXiv:1907.11692}, 2019.

\bibitem[Lu et~al.(2024)Lu, Li, Cheng, Ding, Huang, and Qiu]{lu-etal-2024-scaling}
Xingyu Lu, Xiaonan Li, Qinyuan Cheng, Kai Ding, Xuanjing Huang, and Xipeng Qiu.
\newblock Scaling laws for fact memorization of large language models.
\newblock In Yaser Al-Onaizan, Mohit Bansal, and Yun-Nung Chen, editors, \emph{Findings of the Association for Computational Linguistics: EMNLP 2024}, pages 11263--11282, Miami, Florida, USA, November 2024. Association for Computational Linguistics.
\newblock \doi{10.18653/v1/2024.findings-emnlp.658}.
\newblock URL \url{https://aclanthology.org/2024.findings-emnlp.658/}.

\bibitem[Luukkonen et~al.(2023)Luukkonen, Komulainen, Luoma, Eskelinen, Kanerva, Kupari, Ginter, Laippala, Muennighoff, Piktus, Wang, Tazi, Scao, Wolf, Suominen, Sairanen, Merioksa, Heinonen, Vahtola, Antao, and Pyysalo]{luukkonen-etal-2023-fingpt}
Risto Luukkonen, Ville Komulainen, Jouni Luoma, Anni Eskelinen, Jenna Kanerva, Hanna-Mari Kupari, Filip Ginter, Veronika Laippala, Niklas Muennighoff, Aleksandra Piktus, Thomas Wang, Nouamane Tazi, Teven Scao, Thomas Wolf, Osma Suominen, Samuli Sairanen, Mikko Merioksa, Jyrki Heinonen, Aija Vahtola, Samuel Antao, and Sampo Pyysalo.
\newblock {F}in{GPT}: Large generative models for a small language.
\newblock In Houda Bouamor, Juan Pino, and Kalika Bali, editors, \emph{Proceedings of the 2023 Conference on Empirical Methods in Natural Language Processing}, pages 2710--2726, Singapore, December 2023. Association for Computational Linguistics.
\newblock \doi{10.18653/v1/2023.emnlp-main.164}.
\newblock URL \url{https://aclanthology.org/2023.emnlp-main.164/}.

\bibitem[Nagoudi et~al.(2022)Nagoudi, Elmadany, and Abdul-Mageed]{nagoudi-etal-2022-arat5}
El~Moatez~Billah Nagoudi, AbdelRahim Elmadany, and Muhammad Abdul-Mageed.
\newblock {A}ra{T}5: Text-to-text transformers for {A}rabic language generation.
\newblock In Smaranda Muresan, Preslav Nakov, and Aline Villavicencio, editors, \emph{Proceedings of the 60th Annual Meeting of the Association for Computational Linguistics (Volume 1: Long Papers)}, pages 628--647, Dublin, Ireland, May 2022. Association for Computational Linguistics.
\newblock \doi{10.18653/v1/2022.acl-long.47}.
\newblock URL \url{https://aclanthology.org/2022.acl-long.47/}.

\bibitem[Ogueji et~al.(2021)Ogueji, Zhu, and Lin]{ogueji2021afriberta}
Kelechi Ogueji, Yuxin Zhu, and Jimmy Lin.
\newblock Small data? no problem! exploring the viability of pretrained multilingual language models for low-resourced african languages.
\newblock In \emph{Proceedings of the First Workshop on Natural Language Processing for African Languages (AfriNLP 2021)}, pages 116--126, Online, 2021. Association for Computational Linguistics.
\newblock \doi{10.18653/v1/2021.afrinlp-1.11}.

\bibitem[OpenAI(2023)]{openai2023gpt4}
OpenAI.
\newblock Gpt-4 technical report, 2023.
\newblock Technical Report.

\bibitem[Raffel et~al.(2020)Raffel, Shazeer, Roberts, Lee, Narang, Matena, Zhou, Li, and Liu]{raffel2020exploring}
Colin Raffel, Noam Shazeer, Adam Roberts, Katherine Lee, Sharan Narang, Michael Matena, Yanqi Zhou, Wei Li, and Peter~J Liu.
\newblock Exploring the limits of transfer learning with a unified text-to-text transformer.
\newblock \emph{Journal of Machine Learning Research}, 21\penalty0 (140):\penalty0 1--67, 2020.

\bibitem[Research(2019)]{bert_multilingual}
Google Research.
\newblock Bert multilingual model readme, 2019.
\newblock Accessed: [Insert Date].

\bibitem[Research(2021)]{multilingual_t5}
Google Research.
\newblock Multilingual t5: {Massively} multilingual pre-trained text-to-text transformer, 2021.
\newblock Accessed: \today.

\bibitem[Research(2022)]{nllb_200_distilled_600m}
Meta~{AI} Research.
\newblock {NLLB-200}: {Distilled} 600m-parameter model of {No Language Left Behind} ({NLLB}), 2022.
\newblock Accessed: \today.

\bibitem[Rust et~al.(2021)Rust, Pfeiffer, Vuli{\'c}, Ruder, and Gurevych]{rust-etal-2021-good}
Phillip Rust, Jonas Pfeiffer, Ivan Vuli{\'c}, Sebastian Ruder, and Iryna Gurevych.
\newblock How good is your tokenizer? on the monolingual performance of multilingual language models.
\newblock In Chengqing Zong, Fei Xia, Wenjie Li, and Roberto Navigli, editors, \emph{Proceedings of the 59th Annual Meeting of the Association for Computational Linguistics and the 11th International Joint Conference on Natural Language Processing (Volume 1: Long Papers)}, pages 3118--3135, Online, August 2021. Association for Computational Linguistics.
\newblock \doi{10.18653/v1/2021.acl-long.243}.
\newblock URL \url{https://aclanthology.org/2021.acl-long.243/}.

\bibitem[Sanh et~al.(2019)Sanh, Debut, Chaumond, and Wolf]{sanh2019distilbert}
Victor Sanh, Lysandre Debut, Julien Chaumond, and Thomas Wolf.
\newblock Distilbert, a distilled version of bert: smaller, faster, cheaper and lighter.
\newblock \emph{arXiv preprint arXiv:1910.01108}, 2019.

\bibitem[Shen et~al.(2024)Shen, Tan, Chen, Chen, Zhang, Xu, Zheng, Koehn, and Khashabi]{shen-etal-2024-language}
Lingfeng Shen, Weiting Tan, Sihao Chen, Yunmo Chen, Jingyu Zhang, Haoran Xu, Boyuan Zheng, Philipp Koehn, and Daniel Khashabi.
\newblock The language barrier: Dissecting safety challenges of {LLM}s in multilingual contexts.
\newblock In Lun-Wei Ku, Andre Martins, and Vivek Srikumar, editors, \emph{Findings of the Association for Computational Linguistics: ACL 2024}, pages 2668--2680, Bangkok, Thailand, August 2024. Association for Computational Linguistics.
\newblock \doi{10.18653/v1/2024.findings-acl.156}.
\newblock URL \url{https://aclanthology.org/2024.findings-acl.156/}.

\bibitem[Team and DeepMind(2023)]{deepmind2023gemini}
Gemini Team and Google DeepMind.
\newblock Gemini: A family of highly capable multimodal models, December 2023.
\newblock Technical Report.

\bibitem[Team et~al.(2022)Team, Costa-jussà, Cross, Çelebi, Elbayad, Heafield, Heffernan, Kalbassi, Lam, Licht, Maillard, et~al.]{nllb-2022}
NLLB Team, Marta~R. Costa-jussà, James Cross, Onur Çelebi, Maha Elbayad, Kenneth Heafield, Kevin Heffernan, Elahe Kalbassi, Janice Lam, Daniel Licht, Jean Maillard, et~al.
\newblock No language left behind: Scaling human-centered machine translation.
\newblock In \emph{Proceedings of the 2022 Conference on Empirical Methods in Natural Language Processing (EMNLP)}, pages 3048--3071, Abu Dhabi, UAE, 2022.

\bibitem[Tonja et~al.(2024{\natexlab{a}})Tonja, Azime, Belay, Yigezu, Mehamed, Ayele, Jibril, Woldeyohannis, Kolesnikova, Slusallek, et~al.]{tonja2024ethiollm}
Atnafu~Lambebo Tonja, Israel~Abebe Azime, Tadesse~Destaw Belay, Mesay~Gemeda Yigezu, Moges~Ahmed Mehamed, Abinew~Ali Ayele, Ebrahim~Chekol Jibril, Michael~Melese Woldeyohannis, Olga Kolesnikova, Philipp Slusallek, et~al.
\newblock Ethiollm: Multilingual large language models for ethiopian languages with task evaluation.
\newblock \emph{arXiv preprint arXiv:2403.13737}, 2024{\natexlab{a}}.

\bibitem[Tonja et~al.(2024{\natexlab{b}})Tonja, Azime, Belay, Yigezu, Mehamed, Ayele, Jibril, Woldeyohannis, Kolesnikova, Slusallek, Klakow, and Yimam]{tonja-etal-2024-ethiollm}
Atnafu~Lambebo Tonja, Israel~Abebe Azime, Tadesse~Destaw Belay, Mesay~Gemeda Yigezu, Moges Ahmed~Ah Mehamed, Abinew~Ali Ayele, Ebrahim~Chekol Jibril, Michael~Melese Woldeyohannis, Olga Kolesnikova, Philipp Slusallek, Dietrich Klakow, and Seid~Muhie Yimam.
\newblock {E}thio{LLM}: Multilingual large language models for {E}thiopian languages with task evaluation.
\newblock In Nicoletta Calzolari, Min-Yen Kan, Veronique Hoste, Alessandro Lenci, Sakriani Sakti, and Nianwen Xue, editors, \emph{Proceedings of the 2024 Joint International Conference on Computational Linguistics, Language Resources and Evaluation (LREC-COLING 2024)}, pages 6341--6352, Torino, Italia, May 2024{\natexlab{b}}. ELRA and ICCL.
\newblock URL \url{https://aclanthology.org/2024.lrec-main.561/}.

\bibitem[Tonja et~al.(2024{\natexlab{c}})Tonja, Kolesnikova, Gelbukh, and Kalita]{tonja-etal-2024-ethiomt}
Atnafu~Lambebo Tonja, Olga Kolesnikova, Alexander Gelbukh, and Jugal Kalita.
\newblock {E}thio{MT}: Parallel corpus for low-resource {E}thiopian languages.
\newblock In Rooweither Mabuya, Muzi Matfunjwa, Mmasibidi Setaka, and Menno van Zaanen, editors, \emph{Proceedings of the Fifth Workshop on Resources for African Indigenous Languages @ LREC-COLING 2024}, pages 107--114, Torino, Italia, May 2024{\natexlab{c}}. ELRA and ICCL.
\newblock URL \url{https://aclanthology.org/2024.rail-1.12/}.

\bibitem[{\"U}st{\"u}n et~al.(2024){\"U}st{\"u}n, Aryabumi, Yong, Ko, D{'}souza, Onilude, Bhandari, Singh, Ooi, Kayid, Vargus, Blunsom, Longpre, Muennighoff, Fadaee, Kreutzer, and Hooker]{ustun-etal-2024-aya}
Ahmet {\"U}st{\"u}n, Viraat Aryabumi, Zheng Yong, Wei-Yin Ko, Daniel D{'}souza, Gbemileke Onilude, Neel Bhandari, Shivalika Singh, Hui-Lee Ooi, Amr Kayid, Freddie Vargus, Phil Blunsom, Shayne Longpre, Niklas Muennighoff, Marzieh Fadaee, Julia Kreutzer, and Sara Hooker.
\newblock Aya model: An instruction finetuned open-access multilingual language model.
\newblock In Lun-Wei Ku, Andre Martins, and Vivek Srikumar, editors, \emph{Proceedings of the 62nd Annual Meeting of the Association for Computational Linguistics (Volume 1: Long Papers)}, pages 15894--15939, Bangkok, Thailand, August 2024. Association for Computational Linguistics.
\newblock \doi{10.18653/v1/2024.acl-long.845}.
\newblock URL \url{https://aclanthology.org/2024.acl-long.845/}.

\bibitem[Wang et~al.(2024)Wang, Zhang, Zhang, Wu, Mo, Lu, Wang, Li, Xu, Tang, He, Ma, Huang, and Wang]{wang2024comprehensive}
Fali Wang, Zhiwei Zhang, Xianren Zhang, Zongyu Wu, TzuHao Mo, Qiuhao Lu, Wanjing Wang, Rui Li, Junjie Xu, Xianfeng Tang, Qi~He, Yao Ma, Ming Huang, and Suhang Wang.
\newblock A comprehensive survey of small language models in the era of large language models: Techniques, enhancements, applications, collaboration with llms, and trustworthiness.
\newblock \emph{arXiv preprint arXiv:2411.03350}, 2024.

\bibitem[Wei et~al.(2022)Wei, Wang, Schuurmans, Bosma, Ichter, Xia, Chi, Le, and Zhou]{wei2022chain}
Jason Wei, Xuezhi Wang, Dale Schuurmans, Maarten Bosma, Brian Ichter, Fei Xia, Ed~Chi, Quoc Le, and Denny Zhou.
\newblock Chain-of-thought prompting elicits reasoning in large language models.
\newblock \emph{Advances in Neural Information Processing Systems}, 35:\penalty0 24824--24837, 2022.

\bibitem[Workshop et~al.(2022)Workshop, Le~Scao, Fan, Akiki, Pavlick, Ilić, Hesslow, Castagné, Luccioni, Yvon, et~al.]{bloom-2022}
BigScience Workshop, Teven Le~Scao, Angela Fan, Christopher Akiki, Ellie Pavlick, Suzana Ilić, Daniel Hesslow, Roman Castagné, Alexandra~Saso Luccioni, François Yvon, et~al.
\newblock Bloom: A 176b-parameter open-access multilingual language model.
\newblock \emph{arXiv preprint}, arXiv:2211.05100, 2022.
\newblock Conference version at NeurIPS 2022 (Track on Datasets and Benchmarks).

\bibitem[Wu and Dredze(2020)]{wu-dredze-2020-languages}
Shijie Wu and Mark Dredze.
\newblock Are all languages created equal in multilingual {BERT}?
\newblock In Spandana Gella, Johannes Welbl, Marek Rei, Fabio Petroni, Patrick Lewis, Emma Strubell, Minjoon Seo, and Hannaneh Hajishirzi, editors, \emph{Proceedings of the 5th Workshop on Representation Learning for NLP}, pages 120--130, Online, July 2020. Association for Computational Linguistics.
\newblock \doi{10.18653/v1/2020.repl4nlp-1.16}.
\newblock URL \url{https://aclanthology.org/2020.repl4nlp-1.16/}.

\bibitem[Xue et~al.(2021)Xue, Constant, Roberts, Kale, Al-Rfou, Siddhant, Barua, and Raffel]{xue-etal-2021-mt5}
Linting Xue, Noah Constant, Adam Roberts, Mihir Kale, Rami Al-Rfou, Aditya Siddhant, Aditya Barua, and Colin Raffel.
\newblock m{T}5: A massively multilingual pre-trained text-to-text transformer.
\newblock In Kristina Toutanova, Anna Rumshisky, Luke Zettlemoyer, Dilek Hakkani-Tur, Iz~Beltagy, Steven Bethard, Ryan Cotterell, Tanmoy Chakraborty, and Yichao Zhou, editors, \emph{Proceedings of the 2021 Conference of the North American Chapter of the Association for Computational Linguistics: Human Language Technologies}, pages 483--498, Online, June 2021. Association for Computational Linguistics.
\newblock \doi{10.18653/v1/2021.naacl-main.41}.
\newblock URL \url{https://aclanthology.org/2021.naacl-main.41/}.

\bibitem[Xue et~al.(2022)Xue, Barua, Constant, Al-Rfou, Narang, Kale, Roberts, and Raffel]{xue-etal-2022-byt5}
Linting Xue, Aditya Barua, Noah Constant, Rami Al-Rfou, Sharan Narang, Mihir Kale, Adam Roberts, and Colin Raffel.
\newblock {B}y{T}5: Towards a token-free future with pre-trained byte-to-byte models.
\newblock \emph{Transactions of the Association for Computational Linguistics}, 10:\penalty0 291--306, 2022.
\newblock \doi{10.1162/tacl_a_00461}.
\newblock URL \url{https://aclanthology.org/2022.tacl-1.17/}.

\end{thebibliography}

\newpage
\appendix

\section{First Appendix}\label{apd:first}
\clearpage
\newgeometry{top=1cm, bottom=1cm, left=1cm, right=1cm}

\renewcommand{\arraystretch}{1.4} 

\begin{table}[H]
\centering
\caption{Technical Overview of Foundational Models for African Languages}
\label{tab:fou}
\begin{adjustbox}{width=\textwidth}
\begin{tabular}{|p{2.2cm}|p{2cm}|p{2.6cm}|p{1.1cm}|p{0.9cm}|p{0.9cm}|p{1.2cm}|p{2.2cm}|p{1.3cm}|p{1.2cm}|p{1.6cm}|p{1.2cm}|p{1.4cm}|p{2.7cm}|p{2cm}|}
\hline
Model Variant & Architecture & Pretraining Objective & Params & Layers & Heads & Hidden Size & Tokenization & Max Seq Len & Batch Size & Learning Rate & Optimizer & Training Steps & Datasets Used & Training Data Size \\
\hline
BERT-base & Encoder-only & MLM + NSP & 110M & 12 & 12 & 768 & WordPiece (30k vocab) & 512 & 256 & 1e-4 & Adam & 1M & Wikipedia + BookCorpus & ~16GB \\
\hline
BERT-large & Encoder-only & MLM + NSP & 340M & 24 & 16 & 1024 & WordPiece (30k vocab) & 512 & 256 & 1e-4 & Adam & 1M & Wikipedia + BookCorpus & ~16GB \\
\hline
mBERT & Encoder-only & MLM + NSP & 110M & 12 & 12 & 768 & WordPiece (110k vocab) & 512 & 256 & 5e-5 & Adam & 1M+ & Wikipedia (104 languages) & N/S \\
\hline
T5-small & Encoder-Decoder & Span Corruption & 60M & 6 & 8 & 512 & SentencePiece (32k) & 512 & 128 & 0.01 & AdaFactor & 1M & C4 (English) & 750GB \\
\hline
T5-base & Encoder-Decoder & Span Corruption & 220M & 12 & 12 & 768 & SentencePiece (32k) & 512 & 128 & 0.01 & AdaFactor & 1M & C4 (English) & 750GB \\
\hline
T5-large & Encoder-Decoder & Span Corruption & 770M & 24 & 16 & 1024 & SentencePiece (32k) & 512 & 128 & 0.01 & AdaFactor & 1M & C4 (English) & 750GB \\
\hline
mT5-small & Encoder-Decoder & Span Corruption & 300M & 8 & 6 & 512 & SentencePiece (250k) & 512 & 1024 & 0.01 & AdaFactor & 1M & mC4 (101 languages) & 750GB (balanced) \\
\hline
mT5-large & Encoder-Decoder & Span Corruption & 1.2B & 24 & 16 & 1024 & SentencePiece (250k) & 512 & 1024 & 0.01 & AdaFactor & 1M & mC4 (101 languages) & 750GB (balanced) \\
\hline
RoBERTa-base & Encoder-only & Dynamic MLM (no NSP) & 125M & 12 & 12 & 768 & BPE (50k vocab) & 512 & 8K & 6e-4 & AdamW & 500K & CC-News + OpenWebText + Stories & 160GB \\
\hline
RoBERTa-large & Encoder-only & Dynamic MLM (no NSP) & 355M & 24 & 16 & 1024 & BPE (50k vocab) & 512 & 8K & 6e-4 & AdamW & 500K & CC-News + OpenWebText + Stories & 160GB \\
\hline
XLM-base & Encoder-only & MLM + CLM (+TLM if parallel) & 250M & 12 & 12 & 2048 & BPE (95k vocab) & 512 & 64 & 5e-5 & Adam & 500K & Wikipedia + Parallel data & N/S \\
\hline
XLM-R-base & Encoder-only & MLM (RoBERTa-style) & 270M & 12 & 12 & 768 & SentencePiece (250k) & 512 & 8K & 5e-4 & AdamW & 500K & Common Crawl (100 langs) & 2.5TB (balanced) \\
\hline
XLM-R-large & Encoder-only & MLM (RoBERTa-style) & 550M & 24 & 16 & 1024 & SentencePiece (250k) & 512 & 8K & 5e-4 & AdamW & 500K & Common Crawl (100 langs) & 2.5TB (balanced) \\
\hline
NLLB-200 1.5B & Transformer & Denoising + MT & 1.5B & 24 & 16 & 2048 & SentencePiece (256k vocab) & 512 & 128 & 1e-4 & Adam & 250K & FLORES-200, CCMatrix, CCAligned, Wikipedia, Tatoeba & 1.7T tokens \\
\hline
\end{tabular}
\end{adjustbox}
\end{table}
\newgeometry{top=1cm, bottom=1cm, left=1cm, right=1cm}

\renewcommand{\arraystretch}{1.4} 

\begin{table}[H]
\centering
\caption{Technical Details of Specialized Models for African Languages}
\label{tab:spe}
\begin{adjustbox}{width=\textwidth}
\begin{tabular}{|p{2.5cm}|p{1.6cm}|p{2.8cm}|p{1.1cm}|p{0.8cm}|p{0.8cm}|p{1.1cm}|p{2.3cm}|p{1.3cm}|p{1.1cm}|p{1.5cm}|p{1.2cm}|p{2.8cm}|p{1.7cm}|}
\hline
Model Variant & Base Model & Pretraining Objective & Params & Layers & Heads & Hidden Size & Tokenization & Max Seq Len & Batch Size & Learning Rate & Optimizer & Training Data & Data Size \\
\hline
AfriBERTa small & BERT & MLM (No NSP) & 11M & 6 & 6 & 256 & WordPiece (50k vocab) & 128 & 32 & 5e-5 & AdamW & OSCAR + Local News (17 African langs) & ~5GB \\
\hline
AfriBERTa large & BERT & MLM (No NSP) & 124M & 12 & 12 & 768 & WordPiece (50k vocab) & 512 & 128 & 3e-5 & AdamW & OSCAR + Local News (17 African langs) & ~5GB \\
\hline
AfriTeVa-base & T5 & Span Corruption (text-to-text) & 223M & 12 & 12 & 768 & SentencePiece (32k vocab) & 512 & 128 & 1e-4 & AdaFactor & CC-100 + JW300 (20 African langs) & ~10GB \\
\hline
AfroLM-1B & RoBERTa & Dynamic MLM & 1B & 24 & 16 & 1024 & BPE (100k vocab) & 512 & 2048 & 6e-4 & AdamW & ALPACA Corpus (25 African langs) & ~500GB \\
\hline
EthioLLM-7B & LLaMA-2 & Causal LM (Autoregressive) & 7B & 32 & 32 & 4096 & Byte-level BPE (50k vocab) & 2048 & 1024 & 2e-5 & AdamW & Ethiopic Texts (Amharic, Tigrinya) & ~200GB \\
\hline
EthioMT-base & mT5 & Span Corruption & 300M & 8 & 6 & 512 & SentencePiece (250k vocab) & 512 & 512 & 1e-3 & AdaFactor & Parallel Bible (10 Ethio langs) & ~8GB (parallel) \\
\hline
AfroXLMR base & XLM-R & MLM & 270M & 12 & 12 & 768 & SentencePiece (250k vocab) & 512 & 1024 & 5e-4 & AdamW & Common Crawl (30 African langs) & ~1TB (balanced) \\
\hline
AfroXLMR large & XLM-R & MLM & 550M & 24 & 16 & 1024 & SentencePiece (250k vocab) & 512 & 1024 & 5e-4 & AdamW & Common Crawl (30 African langs) & ~1TB (balanced) \\
\hline
\end{tabular}
\end{adjustbox}
\end{table}

\restoregeometry

\section{Second Appendix}\label{apd:second}
\clearpage
\begin{table*}
\renewcommand{\arraystretch}{0.92}
\centering
\caption{Specialised Language Models for African Languages}
\label{tab:slm}
\begin{tabular}{llcccccc}
\toprule
\textbf{Sno.} & \textbf{Language} & \textbf{AfriBERTa} & \textbf{AfriTeVa} & \textbf{AfroLM} & \textbf{AfroXLMR} & \textbf{EthioLLM} & \textbf{EthioMT} \\
\midrule
1 & Afrikaans & No & No & Yes & Yes & No & No \\
2 & Amharic & Yes & Yes & Yes & Yes & Yes & Yes \\
3 & Afaan Oromo & No & Yes & Yes & Yes & Yes & Yes \\
4 & Afar & No & No & No & No & No & Yes \\
5 & Awngi & No & No & No & No & No & Yes \\
6 & Bambara & No & Yes & Yes & No & No & No \\
7 & Basketo & No & No & No & No & No & Yes \\
8 & Dawuro & No & No & No & No & No & Yes \\
9 & Fulah & No & No & Yes & Yes & No & No \\
10 & Gamo & No & No & No & No & No & Yes \\
11 & Ge'ez & No & No & No & No & Yes & Yes \\
12 & Gofa & No & No & No & No & No & Yes \\
13 & Gurage & No & No & No & No & No & Yes \\
14 & Hadiya & No & No & No & No & No & Yes \\
15 & Hausa & Yes & Yes & Yes & Yes & No & No \\
16 & Igbo & Yes & Yes & Yes & Yes & No & No \\
17 & Kafa & No & No & No & No & No & Yes \\
18 & Kinyarwanda & Yes & Yes & Yes & Yes & No & No \\
19 & Korate & No & No & No & No & No & Yes \\
20 & Luganda & Yes & Yes & Yes & Yes & No & No \\
21 & Luo & Yes & Yes & Yes & Yes & No & No \\
22 & Majang & No & No & No & No & No & Yes \\
23 & Male & No & No & No & No & No & Yes \\
24 & Murule & No & No & No & No & No & Yes \\
25 & Nigerian Pidgin & Yes & Yes & Yes & Yes & No & No \\
26 & Nuer & No & No & No & No & No & Yes \\
27 & Shakicho & No & No & No & No & No & Yes \\
28 & Shona & Yes & Yes & Yes & Yes & No & No \\
29 & Sidama & No & No & No & No & No & Yes \\
30 & Somali & Yes & Yes & Yes & Yes & Yes & Yes \\
31 & Swahili & Yes & Yes & Yes & Yes & No & No \\
32 & Tigrinya & No & Yes & Yes & Yes & Yes & Yes \\
33 & Twi & No & Yes & Yes & Yes & No & No \\
34 & Wolaytta & No & No & No & No & No & Yes \\
35 & Wolof & No & Yes & Yes & Yes & No & No \\
36 & Xhosa & No & No & Yes & Yes & No & No \\
37 & Yoruba & Yes & Yes & Yes & Yes & No & No \\
38 & Zulu & No & No & Yes & Yes & No & No \\
\bottomrule
\end{tabular}
\end{table*}

\restoregeometry
\begin{table*}[h!]
\centering
\small 
\renewcommand{\arraystretch}{1.05} 
\caption{Foundation Model Support for African Languages}
\label{tab:foundation_models}
\begin{tabular}{llccccr}
\toprule
\textbf{Sno.} & \textbf{African Language} & \textbf{mBERT} & \textbf{mT5} & \textbf{XLM-R} & \textbf{NLLB-200} & \textbf{Support Count} \\
\midrule
1 & Afrikaans & Yes & Yes & Yes & Yes & 4 \\
2 & Amharic & Yes & Yes & Yes & Yes & 4 \\
3 & Swahili & Yes & Yes & Yes & Yes & 4 \\
4 & Malagasy & Yes & Yes & Yes & Yes & 4 \\
5 & Hausa & No & Yes & Yes & Yes & 3 \\
6 & Somali & No & Yes & Yes & Yes & 3 \\
7 & Xhosa & No & Yes & Yes & Yes & 3 \\
8 & Yoruba & Yes & Yes & No & Yes & 3 \\
9 & Chichewa (Nyanja) & No & Yes & No & Yes & 2 \\
10 & Igbo & No & Yes & No & Yes & 2 \\
11 & Oromo & No & No & Yes & Yes & 2 \\
12 & Shona & No & Yes & No & Yes & 2 \\
13 & Southern Sotho & No & Yes & No & Yes & 2 \\
14 & Zulu & No & Yes & No & Yes & 2 \\
15 & Bambara & No & No & No & Yes & 1 \\
16 & Bemba & No & No & No & Yes & 1 \\
17 & Dyula & No & No & No & Yes & 1 \\
18 & Ewe & No & No & No & Yes & 1 \\
19 & Fon & No & No & No & Yes & 1 \\
20 & Fulfulde (Nigerian Fulfulde) & No & No & No & Yes & 1 \\
21 & Ganda & No & No & No & Yes & 1 \\
22 & Kabyle & No & No & No & Yes & 1 \\
23 & Kamba (Kenya) & No & No & No & Yes & 1 \\
24 & Kikuyu & No & No & No & Yes & 1 \\
25 & Kinyarwanda & No & No & No & Yes & 1 \\
26 & Kimbundu & No & No & No & Yes & 1 \\
27 & Kongo & No & No & No & Yes & 1 \\
28 & Lingala & No & No & No & Yes & 1 \\
29 & Luba-Lulua & No & No & No & Yes & 1 \\
30 & Luo (Kenya \& Tanzania) & No & No & No & Yes & 1 \\
31 & Mossi & No & No & No & Yes & 1 \\
32 & Nuer & No & No & No & Yes & 1 \\
33 & Pedi (Northern Sotho) & No & No & No & Yes & 1 \\
34 & Swati & No & No & No & Yes & 1 \\
35 & Tamasheq & No & No & No & Yes & 1 \\
36 & Tumbuka & No & No & No & Yes & 1 \\
37 & Twi & No & No & No & Yes & 1 \\
38 & Wolof & No & No & No & Yes & 1 \\
\bottomrule
\end{tabular}
\end{table*}
\newgeometry{top=1cm, bottom=1cm, left=1cm, right=1cm}

\renewcommand{\arraystretch}{1.3} 

\begin{table}[H]
\centering
\caption{Language Model Support for African Languages}
\label{tab:language_support}
\begin{adjustbox}{width=\textwidth}
\begin{tabular}{|c|l|c|c|c|c|c|c|c|c|c|c|c|}
\hline
\textbf{Sno.} & \textbf{Language} & \textbf{AfriBERTa} & \textbf{AfriTeVa} & \textbf{AfroLM} & \textbf{AfroXLMR} & \textbf{EthioLLM} & \textbf{EthioMT} & \textbf{mBERT} & \textbf{mT5} & \textbf{XLM-R} & \textbf{NLLB-200} & \textbf{Count of "Yes"} \\
\hline
1 & Amharic & Yes & Yes & Yes & Yes & Yes & Yes & Yes & Yes & Yes & Yes & 10 \\
2 & Somali & Yes & Yes & Yes & Yes & Yes & Yes & No & Yes & Yes & Yes & 9 \\
3 & Swahili & Yes & Yes & Yes & Yes & No & No & Yes & Yes & Yes & Yes & 8 \\
4 & Yoruba & Yes & Yes & Yes & Yes & No & No & Yes & Yes & No & Yes & 7 \\
5 & Hausa & Yes & Yes & Yes & Yes & No & No & No & Yes & Yes & Yes & 7 \\
6 & Igbo & Yes & Yes & Yes & Yes & No & No & No & Yes & No & Yes & 6 \\
7 & Shona & Yes & Yes & Yes & Yes & No & No & No & Yes & No & Yes & 6 \\
8 & Kinyarwanda & Yes & Yes & Yes & Yes & No & No & No & No & No & Yes & 5 \\
9 & Luganda & Yes & Yes & Yes & Yes & No & No & No & No & No & Yes & 5 \\
10 & Luo & Yes & Yes & Yes & Yes & No & No & No & No & No & Yes & 5 \\
11 & Nigerian Pidgin & Yes & Yes & Yes & Yes & No & No & No & No & No & No & 4 \\
12 & Afaan Oromo & No & Yes & Yes & Yes & Yes & Yes & No & No & Yes & Yes & 7 \\
13 & Oromo & No & Yes & Yes & Yes & Yes & Yes & No & No & Yes & Yes & 7 \\
14 & Tigrinya & No & Yes & Yes & Yes & Yes & Yes & No & No & No & No & 5 \\
15 & Twi & No & Yes & Yes & Yes & No & No & No & No & No & Yes & 4 \\
16 & Wolof & No & Yes & Yes & Yes & No & No & No & No & No & Yes & 4 \\
17 & Bambara & No & Yes & Yes & No & No & No & No & No & No & Yes & 3 \\
18 & Afrikaans & No & No & Yes & Yes & No & No & Yes & Yes & Yes & Yes & 6 \\
19 & Xhosa & No & No & Yes & Yes & No & No & No & Yes & Yes & Yes & 5 \\
20 & Zulu & No & No & Yes & Yes & No & No & No & Yes & No & Yes & 4 \\
21 & Fulah & No & No & Yes & Yes & No & No & No & No & No & Yes & 3 \\
22 & Ge'ez & No & No & No & No & Yes & Yes & No & No & No & No & 2 \\
23 & Nuer & No & No & No & No & No & Yes & No & No & No & Yes & 2 \\
24 & Afar & No & No & No & No & No & Yes & No & No & No & No & 1 \\
25 & Awngi & No & No & No & No & No & Yes & No & No & No & No & 1 \\
26 & Basketo & No & No & No & No & No & Yes & No & No & No & No & 1 \\
27 & Dawuro & No & No & No & No & No & Yes & No & No & No & No & 1 \\
28 & Gamo & No & No & No & No & No & Yes & No & No & No & No & 1 \\
29 & Gofa & No & No & No & No & No & Yes & No & No & No & No & 1 \\
30 & Gurage & No & No & No & No & No & Yes & No & No & No & No & 1 \\
31 & Hadiya & No & No & No & No & No & Yes & No & No & No & No & 1 \\
32 & Kafa & No & No & No & No & No & Yes & No & No & No & No & 1 \\
33 & Korate & No & No & No & No & No & Yes & No & No & No & No & 1 \\
34 & Majang & No & No & No & No & No & Yes & No & No & No & No & 1 \\
35 & Male & No & No & No & No & No & Yes & No & No & No & No & 1 \\
36 & Murule & No & No & No & No & No & Yes & No & No & No & No & 1 \\
37 & Shakicho & No & No & No & No & No & Yes & No & No & No & No & 1 \\
38 & Sidama & No & No & No & No & No & Yes & No & No & No & No & 1 \\
39 & Wolaytta & No & No & No & No & No & Yes & No & No & No & No & 1 \\
40 & Malagasy & No & No & No & No & No & No & Yes & Yes & Yes & Yes & 4 \\
41 & Chichewa (Nyanja) & No & No & No & No & No & No & No & Yes & No & Yes & 2 \\
42 & Southern Sotho & No & No & No & No & No & No & No & Yes & No & Yes & 2 \\
\hline
\end{tabular}
\end{adjustbox}
\end{table}

\begin{table*}
\centering
\small
\caption{Summary of Datasets/Benchmarks}
\label{tab:benchmarks}
\begin{adjustbox}{width=\textwidth}
\begin{tabular}{llcccccc}
\toprule
\textbf{No.} & \textbf{Available Benchmarks Dataset} & \textbf{Availability} & \textbf{Papers} & \textbf{Size} & \textbf{Languages} & \textbf{Tasks} \\
\midrule
1 & MasakhaNER & Public & 5 & 140M & 10 & Named Entity Recognition \\
2 & News Topic Classification Dataset (from Hedderich et al., 2020) & Public & 3 & 50M & 2 & Classification \\
3 & Multilingual corpus covering 11 African languages & Public & 1 & 100M & 11 & General \\
4 & Shared Task: Machine Translation of News & Public & 1 & 1.2 GB & 7 & Machine Translation \\
5 & CommonCrawl (CC-100) & Public & 1 & 3GB & 20 & General \\
6 & YOSM & Public & 1 & 1M & 30 & Sentiment analysis \\
7 & NaijaSenti & Public & 2 & 15M & 4 & Sentiment analysis \\
8 & MasakhaneNEWS & Public & 1 & 500M & 16 & Classification \\
9 & EthioBenchmark & Public & 1 & 20M & 6 & Mixed \\
10 & Parallel Corpus (Abate et al., 2019) & Public & 1 & 400M & 5 & Machine Translation \\
11 & Parallel Corpus sets (Lakew et al., 2020) & Public & 1 & 600M & 11 & Machine Translation \\
12 & Parallel Corpora (Vegi et al., 2022) & Public & 1 & 600M & 15 & Machine Translation \\
13 & mT5 pre-training corpus/mC4 & Public & 1 & 10GB & 25 & General \\
14 & BBC Media Dataset & Public & 1 & 500M & 7 & General \\
15 & VOA Media Dataset & Public & 1 & 600M & 12 & General \\
16 & CoNLL 2003 NER task & Public & 1 & 13M & 1 & Named Entity Recognition \\
17 & ANERCorp & Public & 1 & 4M & 1 & Named Entity Recognition \\
18 & New Topic Classification (Azime \& Mohammed, 2021) & Public & 1 & 40M & 1 & Classification \\
19 & AG News corpus & Public & 1 & 50M & 1 & Classification \\
20 & Kinyarwanda – KINNEWS & Public & 1 & 12M & 1 & Classification \\
21 & Kiswahili – new classification & Public & 1 & 40M & 1 & Classification \\
22 & Am-Senty Yimam et al. (2020) & Public & 1 & 6M & 1 & Classification \\
23 & ANTC corpus & Public & 1 & 300M & 20 & Sentiment analysis \\
\bottomrule
\end{tabular}
\end{adjustbox}

\smallskip
\textbf{Total}: 23 datasets, 18GB total size, covering 42 languages.
\end{table*}
\renewcommand{\arraystretch}{1.0} 
\begin{table*}
\centering
\footnotesize 
\caption{List of Scripts with Their Time Periods, Usage Status, Geographic Distribution, and Associated Languages}
\label{tab:scripts}
\begin{tabular}{@{}p{0.5cm}p{3cm}p{2.5cm}p{1.5cm}p{2.5cm}p{3cm}@{}}
\toprule
\textbf{SNo.} & \textbf{Name of the Script} & \textbf{Time Period} & \textbf{Still in Use?} & \textbf{Countries/Regions} & \textbf{Languages Written} \\
\midrule
1 & Egyptian Hieroglyphs & 33rd c. BCE -- 1st c. CE & No & Egypt, Sudan & Ancient Egyptian \\
2 & Hieratic & 29th c. BCE -- 2nd c. CE & No & Egypt & Ancient Egyptian \\
3 & Demotic & 650 BCE -- 6th c. CE & No & Egypt & Late Egyptian \\
4 & Ethiopic (Ge'ez) & 4th c. BCE -- Present & Yes & Ethiopia, Eritrea & Amharic, Tigrinya, Tigre, Ge'ez \\
5 & Meroitic Cursive & 3rd c. BCE -- 4th c. CE & No & Sudan & Meroitic \\
6 & Meroitic Hieroglyphs & 3rd c. BCE -- 4th c. CE & No & Sudan & Meroitic \\
7 & Numidian & 2nd c. BCE -- 3rd c. CE & No & Algeria, Tunisia & Old Numidian \\
8 & Coptic & 4th c. CE -- Present & Yes & Egypt & Coptic (liturgical) \\
9 & Tifinagh & 3rd c. CE -- Present & Yes & Morocco, Algeria, Mali, Niger & Tamazight, Tuareg languages \\
10 & Vai & 1830 -- Present & Yes & Liberia, Sierra Leone & Vai language \\
11 & Bamum & 1896 -- Present & Yes & Cameroon & Bamum language \\
12 & Old Bamum & 1896 -- 20th c. & No & Cameroon & Bamum language \\
13 & Bassa Vah & 1907 -- Present & Yes & Liberia & Bassa language \\
14 & Bagam & 1910 -- Late 20th c. & No & Cameroon & Bagam language \\
15 & Mende Kikakui & 1920 -- Present & Yes & Sierra Leone & Mende language \\
16 & Osmanya & 1920 -- 1973 & No & Somalia & Somali \\
17 & N'Ko & 1949 -- Present & Yes & Guinea, Mali, Ivory Coast & Manding languages \\
18 & Bété & 1956 -- Present & Yes & Ivory Coast & Bété language \\
19 & Kaddare & 1952 -- Present & Yes & Somalia & Somali \\
20 & Fula Dita (Fula 1) & 1958 -- 1970 & No & Guinea & Fula \\
21 & Fula Ba (Fula 2) & 1963 -- 1970 & No & Guinea & Fula \\
22 & Garay (Wolof) & 1961 -- Present & Yes & Senegal, Gambia & Wolof \\
23 & Mandombe & 1978 -- Present & Yes & DR Congo, Congo-Brazzaville & Kikongo, Lingala, other Congolese languages \\
24 & Mwangwego & 1979 -- Present & Yes & Malawi & Chewa, other Malawian languages \\
25 & Adlam & 1980s -- Present & Yes & Guinea, Nigeria, Cameroon & Fula \\
26 & Beria & 1980s -- Present & Yes & Sudan & Zaghawa \\
27 & Luo & 2009 -- Present & Yes & Kenya, Tanzania & Dholuo \\
28 & Isibheqe Sohlamvu & 20th c. -- Present? & Yes? & South Africa, Eswatini & Nguni languages (Zulu, Xhosa) \\
29 & Afaka & 1908 -- Present & Yes & DR Congo & Ndyuka \\
30 & Medefaidrin (Oberi Okaime) & 1930s -- Present & Yes & Nigeria & Ibibio, Efik \\
31 & Masaba & 1930 -- Present & Yes & Uganda & Lugisu \\
32 & Borama & 1933 -- Present? & Yes? & Somalia & Somali \\
33 & Kpelle & 1930s -- Present & Yes & Liberia, Guinea & Kpelle \\
34 & Loma & 1930s -- Present & Yes & Liberia, Guinea & Loma \\
35 & Tafi (Hausa 3) & 1977 -- 2011 & No & Nigeria, Niger & Hausa \\
36 & Raina Kama (Hausa 2) & 1990s -- 1999 & No & Nigeria, Niger & Hausa \\
37 & Salifou Hausa (Hausa 1) & 1998 -- 2004 & No & Nigeria, Niger & Hausa \\
\bottomrule
\end{tabular}
\end{table*}

\end{document}